\documentclass{article} % For LaTeX2e

\usepackage[nonatbib,final]{neurips_2022}
\usepackage[numbers]{natbib}

\usepackage{amsmath,amsfonts,bm}

\def\eqref#1{equation~\ref{#1}}
\def\1{\bm{1}}

\def\vb{{\bm{b}}}

\def\ve{{\bm{e}}}

\def\vh{{\bm{h}}}

\def\vr{{\bm{r}}}

\def\vt{{\bm{t}}}

\def\vx{{\bm{x}}}
\def\vy{{\bm{y}}}

\def\mE{{\bm{E}}}

\def\mM{{\bm{M}}}

\def\mR{{\bm{R}}}

\def\mW{{\bm{W}}}

\DeclareMathAlphabet{\mathsfit}{\encodingdefault}{\sfdefault}{m}{sl}
\SetMathAlphabet{\mathsfit}{bold}{\encodingdefault}{\sfdefault}{bx}{n}

\def\gA{{\mathcal{A}}}

\def\gC{{\mathcal{C}}}
\def\gD{{\mathcal{D}}}
\def\gE{{\mathcal{E}}}

\def\gG{{\mathcal{G}}}

\def\gN{{\mathcal{N}}}

\def\gQ{{\mathcal{Q}}}
\def\gR{{\mathcal{R}}}

\def\gT{{\mathcal{T}}}

\def\gV{{\mathcal{V}}}

\def\sR{{\mathbb{R}}}

\usepackage{hyperref}
\usepackage{url}

\usepackage{subfig}
\usepackage[utf8]{inputenc} % allow utf-8 input
\usepackage[T1]{fontenc}    % use 8-bit T1 fonts

\usepackage{tikz}
\usetikzlibrary{arrows.meta, bending, calc, chains, positioning}
\usepackage{array,booktabs}       % professional-quality tables
\usepackage{multirow}
\usepackage{amsfonts}       % blackboard math symbols
\usepackage{nicefrac}       % compact symbols for 1/2, etc.
\usepackage{microtype}      % microtypography
\usepackage{pgfplots}
\pgfplotsset{compat=1.9}
\usepackage{xcolor}         % colors
\usepackage{graphicx}
\usepackage{amsthm}
\usepackage{wrapfig}
\usepackage{adjustbox}
\usepackage{bm}

\hypersetup{
    colorlinks,
    linkcolor={red!50!black},
    citecolor={blue!50!black},
    urlcolor={blue!80!black}
}

\definecolor{picred}{RGB}{204, 0, 0}
\definecolor{picyel}{RGB}{241, 194, 50}
\definecolor{picgreen}{RGB}{106, 168, 79}
\definecolor{cerulean}{rgb}{0.0, 0.48, 0.65}

\title{Inductive Logical Query Answering \\ in Knowledge Graphs}

\author{%
   Mikhail Galkin \\%\thanks{Correspondence to Mikhail Galkin \texttt{mikhail.galkin@mila.quebec} or Jian Tang \texttt{jian.tang@hec.ca}} \\
   Mila, McGill University \\
   \texttt{mikhail.galkin@mila.quebec} \\
   \And
   Zhaocheng Zhu \\
   Mila, Universit\'e de Montr\'eal \\
   \texttt{zhuzhaoc@mila.quebec} \\
   \AND
   Hongyu Ren \\
   Stanford University \\
   \texttt{hyren@stanford.edu} \\
   \And
   Jian Tang \\
   Mila, HEC Montr\'eal, CIFAR AI Chair \\
   \texttt{jian.tang@hec.ca} \\
}

\begin{document}

\newcommand{\mycomment}[3]{\textcolor{#1}{[\bf #2: #3]}}
\newcommand{\jt}[1]{\mycomment{orange}{Jian}{#1}}
\newcommand{\zh}[1]{\mycomment{magenta}{Zhaocheng}{#1}}
\newcommand{\mg}[1]{\mycomment{green!70!black}{Mike}{#1}}
\newcommand{\inlinegraphics}[1]{
  \begingroup\normalfont
  \includegraphics[height=1.1\fontcharht\font`\B]{#1}
  \endgroup
}
\newcommand{\edit}[1]{\textcolor{cerulean}{#1}}

\maketitle
\begin{abstract}

% main communication interface
Formulating and answering logical queries is a standard communication interface for knowledge graphs (KGs). % and their representations. 
Alleviating the notorious incompleteness of real-world KGs, neural methods achieved impressive results in link prediction and complex query answering tasks by learning representations of entities, relations, and queries.
% problems: impossible to answer queries over new graphs and unseen entities, generalization to larger graphs
Still, most existing query answering methods rely on transductive entity embeddings %are inherently transductive 
and cannot generalize to KGs containing new entities without retraining the entity embeddings. 
% approach: inference-only (CQD) and learning (GNN-QE)
In this work, we study the inductive query answering task where inference is performed on a graph containing new entities with queries over both seen and unseen entities.
%To this end, we devise two mechanisms leveraging graph neural networks and recently proposed KG representation models with inductive capabilities.
To this end, we devise two mechanisms leveraging inductive \emph{node} and \emph{relational structure} representations powered by graph neural networks (GNNs).
% contributions: inductive query answering, new datasets
% exps: inductive QA, generalization to larger graphs (20-400% larger)
%Experimentally, we show that message passing GNNs are a key component in the inductive setup generalizing to unseen nodes and query patterns. \zh{I think our biggest contribution is about the understanding of inductiveness. Message passing GNNs is just a way to instantiate it, and not all message passing GNNs are inductive.}
%Varying the ratio of new nodes at inference time, GNNs retain the performance on graphs up to 300\% larger than training ones.
Experimentally, we show that inductive models are able to perform logical reasoning at inference time over unseen nodes generalizing to graphs up to 500\% larger than training ones. 
Exploring the efficiency--effectiveness trade-off, we find the inductive \emph{relational structure} representation method generally achieves higher performance, while the inductive \emph{node representation} method is able to answer complex queries in the \emph{inference-only} regime without any training on queries and scales to graphs of millions of nodes. 
Code is available at \url{https://github.com/DeepGraphLearning/InductiveQE}.
\end{abstract}

\section{Introduction}

Traditionally, querying knowledge graphs (KGs) is performed via databases using structured query languages like SPARQL. 
Databases can answer complex queries relatively fast under the assumption of \emph{completeness}, i.e., there is no missing information in the graph. 
In practice, however, KGs are notoriously incomplete~\citep{DBLP:conf/www/WestGMSGL14}.
Embedding-based methods that learn vector representations of entities and relations are known to be effective in \emph{simple link prediction} predicting heads or tails of query patterns $\textit{(head, relation, ?)}$, e.g., \emph{(Einstein, graduate, ?}), as common in 
%the field of 
\emph{KG completion}~\citep{ali2020light, ji2022kgs}.

\begin{figure}[t]
    \centering
    \includegraphics[width=\textwidth]{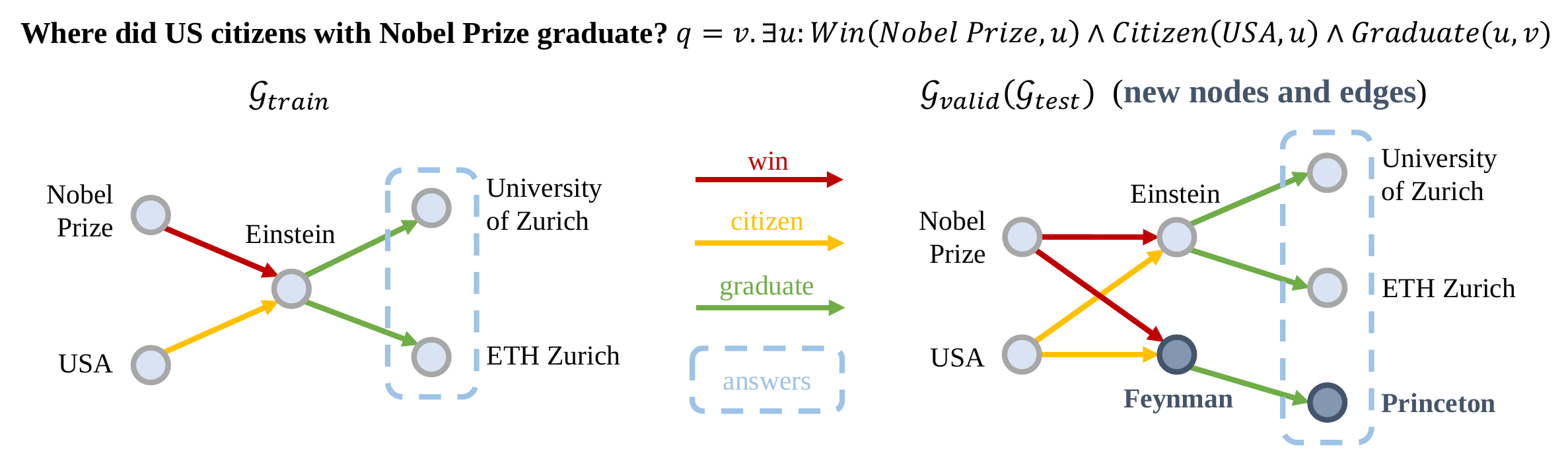}
    \caption{Inductive query answering problem: at inference time, the graph is updated with new nodes \texttt{Feynman} and \texttt{Princeton} and edges such that the same query now has more answers.}
    \label{fig:intro1}
\end{figure}

Complex queries are graph patterns expressed in a subset of first-order logic (FOL) with operators such as intersection ($\land$), union ($\lor$), negation ($\lnot$) and existentially quantified ($\exists$) variables\footnote{The universal quantifier ($\forall$) is often discarded as in real-world KGs there is no node connected to all others.}, e.g., $?U.\exists V : \texttt{Win}(\texttt{NobelPrize}, V) \land \texttt{Citizen}(\texttt{USA}, V) \land \texttt{Graduate}(V, U)$ (Fig.~\ref{fig:intro1}).
%\jt{I would still suggest to give an example here to make the paper self-contained.} 
% Graph patterns consist of \emph{atomic} triple patterns like $\texttt{Citizen}(\texttt{Canada}, V)$ %$\textit{(head, relation, ?)}$
% such that simple link prediction models a \emph{projection} operator and can be considered as a subset of complex query answering task.
%$\textit{(head, relation, ?)}$
Complex queries define a superset of KG completion. The conventional KG completion (link prediction) task can be viewed as a complex query with a single triplet pattern without logical operators, e.g., $\texttt{Citizen}(\texttt{USA}, V)$, which we also denote as a \emph{projection} query. 

To tackle complex queries on incomplete knowledge graphs, \emph{query embedding} methods are proposed to execute logic operations in the latent space, including variants
% Executing such logical operators in latent space has led to the explosion of \emph{query embedding} (QE) methods
that employ geometric~\citep{hamilton2018embedding,ren2019query2box,zhang2021cone}, probabilistic~\citep{ren2020beta, choudhary2021probabilistic}, neural-symbolic~\citep{DBLP:conf/nips/SunAB0C20,chen2021fuzzy,arakelyan2020complex}, neural~\citep{biqe,amayuelas2022neural}, and GNN~\citep{daza2020message,alivanistos2022query} approaches for learning entity, relation, and query representations. 

However, this very fact of learning a separate embedding for each entity makes those methods inherently \emph{transductive}
i.e., they are bound to the space of learned entities and cannot generalize to % new KGs or new
unseen entities without retraining the whole embedding matrix which can be prohibitively expensive in large graphs. %In contrast, \emph{inductive} methods are able to 
The problem is illustrated in Fig.~\ref{fig:intro1}: given a graph about \texttt{Einstein} and a logical query \emph{Where did US citizens with Nobel Prize graduate?}, transductive QE methods learn to execute logical operators and return the answer set $\{ \texttt{University of Zurich, ETH Zurich} \}$. 
Then, the graph is updated with new nodes and edges about \texttt{Feynman} and \texttt{Princeton}, %\zh{Does it break the anonymous rules?}
and the same query now has more correct answers $\{ \texttt{University of Zurich, ETH Zurich, Princeton} \}$ as new unseen entities satisfy the query as well.

Such \emph{inductive inference} is not possible for transductive models as they do not have representations for new \texttt{Feynman} and \texttt{Princeton} nodes. In the extreme case, inference graphs might be disconnected from the training one and only share the set of relations.
Therefore, inductive capabilities are a key factor to % enable
transferring trained query answering models onto updated or entirely new KGs.

%\jt{I would suggest to reorganize the contributions of this paper: (1) This paper is the first work that studied complex logical query answering in the inductive setting. We constructed a new set of data sets for this tasks, which are able to serve as the benchmark data for this task in the future; (2) We proposed two mechanisms based on... (3) Experimental results ..., emphasizing out-of-distribution generalization (generalize to much larger graphs with new entities.}
In this work, we study answering complex queries in the inductive setting, where the model has to deal with unseen entities at inference time. 
Inspired by recent advancement in inductive Kg completion~\citep{zhu2021neural, galkin2022nodepiece}, we devise two solutions for learning inductive representations for complex query: 1) The first solution, NodePiece-QE, extends the inductive node representation model NodePiece~\cite{galkin2022nodepiece} to complex query answering. NodePiece-QE learns \textbf{inductive representations of each entity} as a function of tokens from a fixed-size vocabulary, and answers complex query with a non-parametric logical query executor~\citep{arakelyan2020complex}. The advantages of NodePiece-QE are that it only needs to be trained on simple link prediction data, answers complex queries in the \emph{inference-only} mode, and % that it
can scale to large KGs. 2) The second solution, GNN-QE~\citep{gnn_qe}, extends the inductive KG completion model NBFNet~\cite{zhu2021neural} for complex query answering. Originally, GNN-QE was studied only in the transductive setting. Here, we analyze its inductive capabilities. GNN-QE learns \textbf{inductive representations of the relational structure} without entity embeddings, and uses the relational structure between the query constants and the answers to make the prediction. GNN-QE can be trained end-to-end on complex queries, achieves much better performance than NodePiece-QE, but struggles to scale to large KGs.

% In this work, we study complex logical query answering in the inductive mode where reasoning at inference time has to be performed over both seen and unseen entities. 
% To deal with representations of unseen entities, we devise two mechanisms based on the recent progress in inductive link prediction: 1) an \emph{inference-only} embedding-based method that bootstraps seen and unseen entity representations from a universal fixed-size vocabulary~\citep{galkin2022nodepiece} followed by a non-parametric logical query executor~\citep{arakelyan2020complex}; and 2) employing a neural-symbolic method~\citep{gnnqe} with an end-to-end \emph{trainable} projection operator that does not need trainable entity embeddings. \jt{It is a little bit difficult for me to understand the motivations of designing such a method according to such a paragraph.}
To the best of our knowledge, this is the first work to study complex logical query answering in the inductive setting \emph{without any additional features like entity types or textual descriptions}. 
%We design a novel benchmarking suite of 10 datasets of various scales and graph characteristics.
Conducting experiments on a novel benchmarking suite of 10 datasets, we find that 1) both inductive solutions exhibit non-trivial performance answering logical queries over unseen entities and query patterns; 2) inductive models demonstrate out-of-distribution generalization capabilities to graphs up to 500\% larger than training ones; 3) akin to updatable databases, inductive methods can successfully find new correct answers to known training queries after adding new nodes and edges; 4) the inductive \emph{node representation} method scales to answering logical queries over a graph of 2M nodes with 500k new unseen nodes; 5) GNN-based models still exhibit some difficulties~\cite{DBLP:conf/nips/KnyazevTA19, DBLP:conf/icml/YehudaiFMCM21} generalizing to graphs larger than those they were originally trained on.

\section{Related Work}

% \textbf{KG completion.}
\textbf{Knowledge Graph Completion.}
% Traditionally,
Knowledge graph completion, a.k.a. simple link prediction, has been widely studied in the \emph{transductive} paradigm~\cite{DBLP:conf/nips/BordesUGWY13, DBLP:journals/corr/YangYHGD14, DBLP:conf/iclr/SunDNT19, DBLP:conf/nips/0007TYL19}, i.e., when training and inference are performed on the same graph with a fixed set of entities. % where learning shallow embedding vector for each node would suffice.
Generally, these methods learn a shallow embedding vector for each entity.
We refer the audience to respective surveys~\citep{ali2020light,ji2022kgs} covering dozens of % proposed models.
transductive embedding methods.
The emergence of message passing~\citep{DBLP:conf/icml/GilmerSRVD17} and graph neural networks (GNNs) has led to more advanced, \emph{inductive} representation learning approaches % without trainable entity embeddings. 
that model entity or triplet representations as a function of the graph structure in its neighborhood.
% There, node representations can be built from neighborhood features~\citep{DBLP:conf/icml/TeruDH20}, initialized with a query relation type~\citep{zhu2021neural}, or composed from a set of incident relation types~\citep{galkin2022nodepiece}. 
GraIL~\cite{DBLP:conf/icml/TeruDH20} learns triplet representations based on the subgraph structure surrounding the two entities. NeuralLP~\cite{DBLP:conf/nips/YangYC17}, DRUM~\cite{sadeghian2019drum} and NBFNet~\cite{zhu2021neural} learn the pairwise entity representations based on the set of relation paths between two entities. NodePiece~\cite{galkin2022nodepiece} learns entity representations from a fixed-size vocabulary of tokens that can be anchor nodes in a graph or relation types.

\textbf{Complex Query Answering.}
In the complex (multi-hop) query answering setup with logical operators, existing models employ different approaches, e.g., geometric~\citep{hamilton2018embedding,ren2019query2box,zhang2021cone}, probabilistic~\citep{ren2020beta, choudhary2021probabilistic}, neural-symbolic~\citep{DBLP:conf/nips/SunAB0C20,chen2021fuzzy,arakelyan2020complex}, neural~\citep{biqe,amayuelas2022neural}, and GNN~\citep{daza2020message,alivanistos2022query}.
Still, all the approaches are created and evaluated exclusively in the transductive mode where the set of entities does not change at inference time.
To the best of our knowledge, there is no related work in inductive logical query answering when the inference graph contains new entities.
With our work, we aim to bridge this gap and extend inductive representation learning algorithms to logical query answering.
In particular, we focus on the inductive setup where an inference graph is a superset of a training graph\footnote{The set of relation types is fixed.} such that 1) inference queries require reasoning over both seen and new entities; 2) original training queries might have more correct answers at inference time with the addition of new entities.

% All methods that learn entity embeddings are transductive. 
% NodePiece and NBFNet allow for transfer to new graphs.
% 2 ways: 1) obtaining embeddings of new nodes + running an inference-only CQD; 2) training a GNN-QE

\section{Preliminaries and Problem Definition}

% MG: Done
% \zh{I feel inductive setup may go to the next section. Preliminary is usually something that people can ignore if they feel they are experts in this field. Here the inductive complex query is totally defined by our paper and should be counted in the methodology section.}

\textbf{Knowledge Graph and Inductive Setup.} 
Given a finite set of entities $\gE$, a finite set of relations $\gR$, and a set of triples (edges) $\gT = (\gE \times \gR \times \gE)$, a knowledge graph $\gG$ is defined as $\gG = (\gE, \gR, \gT)$. Accounting for the inductive setup, we define a \emph{training} graph $\gG_{\textit{train}} = (\gE_{\textit{train}}, \gR, \gT_{\textit{train}})$ and an \emph{inference} graph $\gG_{\textit{inf}} = (\gE_{\textit{inf}}, \gR, \gT_{\textit{inf}})$ such that $\gE_{\textit{train}} \subset \gE_{\textit{inf}}$ and $\gT_{\textit{train}} \subset \gT_{\textit{inf}}$. That is, the \emph{inference} graph extends the training graph with new entities and edges\footnote{Note that the set of relation types $\gR$ remains the same.}.% \jt{why do we have to have these two constraints?$\gE_{\textit{train}} \subset \gE_{\textit{inf}}$ and $\gT_{\textit{train}} \subset \gT_{\textit{inf}}$. I think a more interesting setting is to simply make sure the set of relations are the same across different graphs in training and inference.}
The inference graph $\gG_{\textit{inf}}$ is an incomplete part of the not observable complete graph $\hat{\gG}_{\textit{inf}} = (\gE_{\textit{inf}}, \gR, \hat{\gT}_{\textit{inf}})$ with $\hat{\gT}_{\textit{inf}} = \gT_{\textit{inf}} \cup \gT_{\textit{pred}}$ whose missing triples $\gT_{\textit{pred}}$ have to be predicted at inference time. %\zh{Personally I feel a paragraph with a lot of inline equations is not very friendly to readers.}

\textbf{First-Order Logic Queries.}
Applied to KGs, a first-order logic (FOL) query $\gQ$ is a formula that consists of constants $\gC$ ($\gC \subseteq \gE$), variables $\gV$ ($\gV \subseteq \gE$, existentially quantified), relation \emph{projections} $R(a, b)$ denoting a binary function over constants or variables, and logic symbols ($\exists, \land, \lor, \lnot$).
The answers $A_{\gG}(\gQ)$ to the query $\gQ$ are assignments of variables in a formula such that the instantiated query formula is a subgraph of the complete graph $\hat{\gG}$.

Fig.~\ref{fig:intro1} illustrates the logical form of a query \emph{Where did US citizens with Nobel Prize graduate?} as $?U.\exists V : \texttt{Win}(\texttt{NobelPrize}, V) \land \texttt{Citizen}(\texttt{USA}, V) \land \texttt{Graduate}(V, U)$ where \texttt{NobelPrize} and \texttt{USA} are \emph{constants}; \texttt{Win}, \texttt{Citizen}, \texttt{Graduate} are \emph{relation projections} (labeled edges); $V, U$ - \emph{variables} such that $V$ is an existentially quantified free variable and $U$ is the projected bound \emph{target} variable of the query.
Common for the literature, we aim at predicting assignments of the query \emph{target} whereas assignments of intermediate variables might not always be explicitly interpreted depending on the model architecture.
In the example, the answer set $A_{\gG}(\gQ)$ is a binding of a target variable $U$ to constants \texttt{University of Zurich} and \texttt{ETH Zurich}.

%\jt{also try to elaborate a little more on why inductive generalization is very important for this task.}
\textbf{Inductive FOL Queries.}
In the standard transductive query answering setup, query constants and variables at both training and inference time belong to the same set of entities, i.e., $\gC_{\textit{train}} = \gC_{\textit{inf}} \subseteq \gE, \gV_{\textit{train}} = \gV_{\textit{inf}} \subseteq \gE$.
% \gC_{\textit{inf}} \subseteq \gE, \gV_{\textit{inf}} \subseteq \gE
In the inductive setup covered in this work, query constants and variables at inference time belong to a different and larger set of entities $\gE_{\textit{inf}}$ from the inference graph $\gG_{\textit{inf}}$, i.e., $\gC_{\textit{train}} \subseteq \gE_{\textit{train}}, \gV_{\textit{train}} \subseteq \gE_{\textit{train}}$ but  $\gC_{\textit{inf}} \subseteq \gE_{\textit{inf}},  \gV_{\textit{inf}} \subseteq \gE_{\textit{inf}}$.
This also leads to the fact that training queries executed over the inference graph might have more correct answers, i.e., $A_{\gG_{\textit{train}}}(\gQ) \subseteq A_{\gG_{\textit{inf}}}(\gQ)$.
For example (cf. Fig.~\ref{fig:intro1}), the inference graph is updated with new nodes \texttt{Feynman}, \texttt{Princeton} and their new respective edges. 
The same query now has a larger set of intermediate variables satisfying the formula (\texttt{Feynman}) and an additional correct answer \texttt{Princeton}.
Therefore, inductive generalization is essential for obtaining representations of such new nodes and enabling logical reasoning over both seen and new nodes, i.e., finding more answers to known queries in larger graphs or answering new queries with new constants. 
In the following section, we describe two approaches for achieving inductive generalization with different parameterization strategies.

\section{Method}

% \zh{
% One way I could think of organizing this section is
% \begin{enumerate}
%     \item introduce the inductive setting of complex query
%     \item discuss the essence of "inductiveness". With what kind of parameterization do we get an inductive model for inference.
%     \item motivate two frameworks. one is to make unseen nodes as a function of existing nodes (NodePiece), one is to consider relative relation structures without entity embeddings (NBFNet). We may illustrate with some math equations.
%     \item technical details of NodePiece + CompGCN
%     \item technical details of GNN-QE
% \end{enumerate}
% Since we never discussed the essence of inductiveness in NodePiece nor GNN-QE, I think this should be regarded as a contribution to the community.
% }

%\jt{I'm not sure I understand the logic in the following paragraph. You can motivate in the following way. To develop an algorithm that works in the inductive setting, i.e., generalize to a graph with new nodes, there are two potential solutions: (1) there is an inductive node representation learning algorithm (corresponds to your NodePiece algorithm); or (2) the reasoning algorithm does not make use of node representations and only relies on relations (NBFNet).}

\textbf{Inductive Representations of Complex Queries.}
Given a complex query $\gQ=(\gC, \gR_\gQ, \gG)$, the goal is to rank all possible entities according to the query. From a representation learning perspective, this requires us to learn a conditional representation function $f(e|\gC, \gR_\gQ, \gG)$ for each entity $e \in \gE$. Transductive methods learn a shallow embedding for each answer entity $e \in \gE$, and, therefore, cannot generalize to unseen entities. For inductive methods, the function $f(e|\gC, \gR_\gQ, \gG)$ should generalize to some unseen answer entity $e'$ (or unseen constant entity $c' \in \gC'$) at inference time. Here, we discuss two solutions for devising such an inductive function.

\begin{figure}[t]
    \centering
    \includegraphics[width=\textwidth]{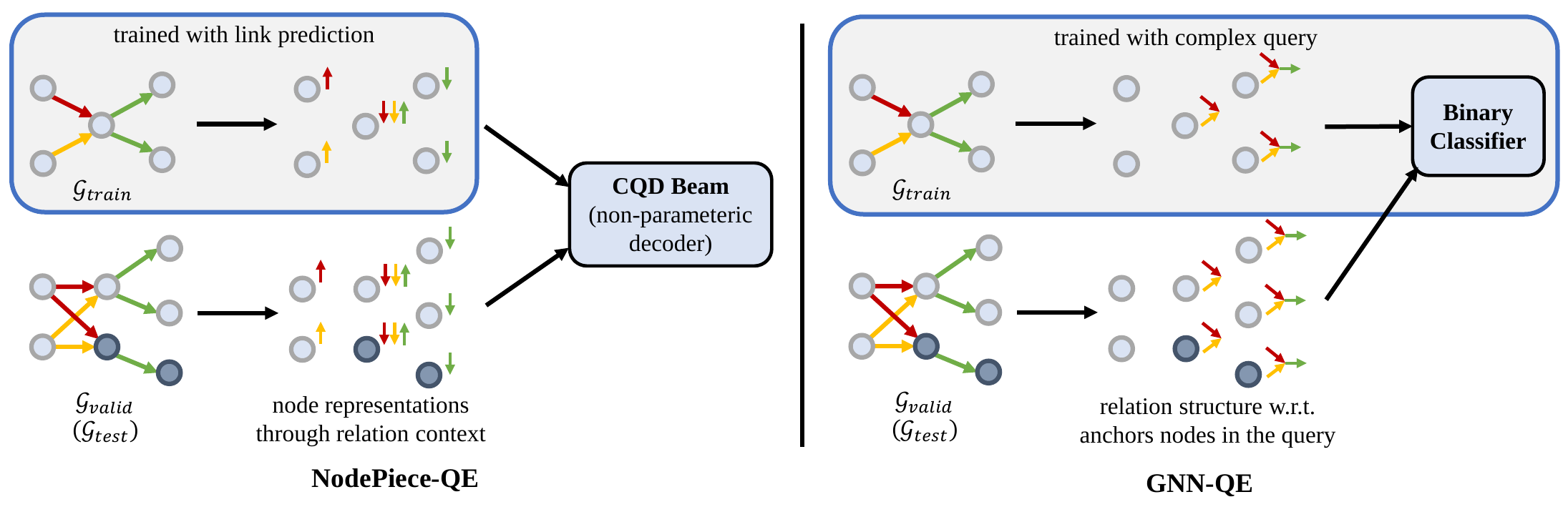}
    \caption{Inductive node representation (NodePiece-QE, left) and relational structure (GNN-QE, right) strategies for complex logical query answering. In NodePiece-QE, we obtain inductive node representations through the invariant set of tokens (here, through incident relation types). NodePiece-QE is an inference-only approach, pre-trained with simple \emph{1p} link prediction and can be directly applied to inductive complex queries with a non-parametric decoder (e.g., CQD Beam). In GNN-QE, we learn the the relative structure of each node w.r.t. the anchor nodes in the query. GNN-QE is trainable end-to-end on \emph{complex queries}.}
    \label{fig:method}
\end{figure}

The first solution is to \textbf{parameterize the representation of each entity $e$ as a function of an invariant vocabulary} of \emph{tokens} that does not change at training and inference. Particularly, the vocabulary might consist of unique relation types $\gR$ that are always the same for $\gG_{\textit{train}}$ and $\gG_{\textit{inf}}$, and we are able to infer the representation of an unseen answer entity (or an unseen constant entity) as a function of its incident relations (cf. Fig.~\ref{fig:method} left). The idea has been studied in NodePiece~\cite{galkin2022nodepiece} for simple link prediction. Here, we adopt a similar idea to learn inductive entity representations for complex query answering. Once we obtain the representations for unseen entities, we can use any off-the-shelf decoding method (e.g., CQD-Beam~\cite{arakelyan2020complex}) for predicting the answer to the complex query. 
We denote this strategy as NodePiece-QE.

The second solution is to \textbf{parameterize $f(e|\gC, \gR_\gQ, \gG)$ as a function of the relational structure}. Intuitively, an answer of a complex query can be decided solely based on the relational structure between the query constants and the answer (Fig.~\ref{fig:intro1}). 
%If we anonymize the name of each entity, we are still able to predict the entity McGill is an answer, since it forms a Y-shape \inlinegraphics{img/Y_arrows.pdf} with the query constants and conforms the query structure. 
Even after anonymizing entity names (and, hence, not learning any explicit entity embedding), we can still infer \texttt{Princeton} as an answer since it forms a distinctive relational structure \inlinegraphics{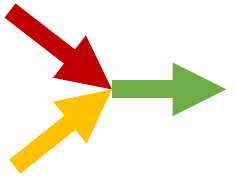} with the query constants and conforms to the query structure. Similarly, intermediate nodes will be deemed correct if they follow a relational structure \inlinegraphics{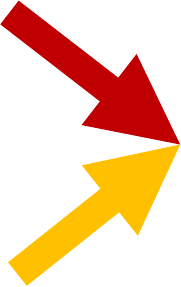}.
In other words, we do not need to know the answer node is \texttt{Princeton}, but only need to know the relative position of \texttt{Princeton} w.r.t. the constants like \texttt{Nobel Prize} and \texttt{USA}. 
Based on this idea, we design $f(e|\gC, \gR_\gQ, \gG)$ to be a relational structure search function. Such an idea has been studied in Neural Bellman-Ford Networks (NBFNet)~\cite{zhu2021neural} to search for a single relation in simple link prediction.
Applied to complex queries, GNN-QE~\cite{gnn_qe} chains several NBFNet instances with differentiable logic operations to learn inductive complex query in an end-to-end fashion.
So far, GNN-QE was evaluated solely on transductive tasks. Here we extend it to the inductive setup.

% \begin{figure}[t]
%     \centering
%     \includegraphics[width=\textwidth]{img/InductiveQA_3.pdf}
%     \caption{}
%     \label{fig:method2}
% \end{figure}

\subsection{NodePiece-QE: Inductive Node Representation}
Here we aim at reconstructing node representations for seen and unseen entities without learning shallow node embedding vectors. 
%\zh{Here, we aim at learning an invariant vocabulary for computing inductive entity representations.}
To this end, we employ NodePiece~\cite{galkin2022nodepiece}, a compositional tokenization approach that learns an invariant vocabulary of \emph{tokens} shared between training and inference graphs. %\zh{that learns the inductive entity representations based on a few entities and relations shared between training and inference graphs}.
Formally, given a vocabulary of tokens $t_i \in T$, each entity $e_i$ is deterministically hashed into a set of representative tokens $e_i = [t_1, \dots, t_k]$. 
An entity vector $\ve_i$ is then obtained as a function of token embeddings $\ve_i = f_{\theta}([\vt_i, \dots, \vt_k]), \vt_i \in \sR^d$ where the encoder function $f_{\theta} : \sR^{k \times d} \rightarrow \sR^d$ is parameterized with a neural network $\theta$.

Since the set of relation types $\gR$ is invariant for training and inference graphs, we can learn relation embeddings $\mR \in \sR^{|\gR|\times d}$ and our vocabulary of learnable tokens $T$ is comprised of distinct relation types such that entities are hashed into a set of unique incident relation types. 
For example (cf. Fig.~\ref{fig:method} left), a middle node from a training graph $\gG_{\textit{train}}$ is hashed with a set of relations $e_i = [\inlinegraphics{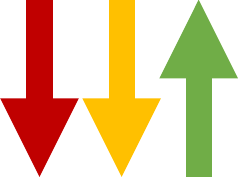}]$ that stands for two unique incoming relations \inlinegraphics{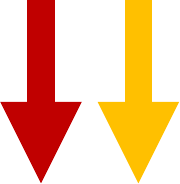} and one unique outgoing relation \inlinegraphics{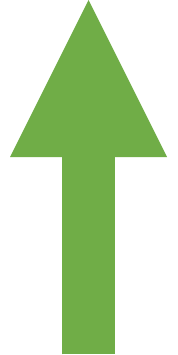}.
Passing the hashes through $f_{\theta}$, we can reconstruct the whole entity embedding matrix $\mE \in \sR^{|\gE_{\textit{train}}| \times d}$.
Additionally, it is possible to enrich entity and relation embeddings by passing them through a relational GNN encoder~\cite{Vashishth2020Composition-based} over a target graph $\gG$: $\mE', \mR' = \text{GNN}(\mE, \mR, \gG)$.
In both ways, the entity embedding matrix $\mE$ encodes a \emph{joint} probability distribution $p(h,r,t)$ for all triples in a graph.

Having a uniform featurization mechanism for both seen and unseen entities, it is now possible to apply any previously-transductive complex query answering model with learnable entity embeddings and logical operators~\cite{ren2019query2box,daza2020message,ren2020beta,chen2021fuzzy}. 
Moreover, it was recently shown~\cite{arakelyan2020complex} that a combination of simple link prediction pre-training and a non-parametric logical executor allows to effectively answer complex FOL queries in the \emph{inference-only} regime without training on any complex query sample.
We adopt this Continuous Query Decomposition algorithm with beam search (CQD-Beam) as the main query answering decoder.
CQD-Beam relies only on entity and relation embeddings $\mE, \mR$ pre-trained on a simple \emph{1p} link prediction task. 
Then, given a complex query, CQD-Beam applies \emph{t-norms} and \emph{t-conorms}~\cite{DBLP:journals/fss/KlementMP04a} that execute conjunctions ($\land$) and disjunctions ($\lor$) as non-parametric algebraic operations in the embedding space, respectively.

In our inductive setup (Fig.~\ref{fig:method}), we train a NodePiece encoder $f_{\theta}$ and relation embeddings $\mR$ (and optionally a GNN) on the \emph{1p} link prediction task over the training graph $\gG_{\textit{train}}$.
We then apply the learned encoder to materialize entity representations of the inference graph $\mE \in \sR^{|\gE_{\textit{inf}}| \times d}$ and send them to CQD-Beam that performs a non-parametric decoding of complex FOL queries over new unseen entities.
The inference-only nature of NodePiece-QE is designed
%to be challenging and probing 
to probe the abilities for zero-shot generalization in performing complex logical reasoning over larger graphs. 
%\mg{Something about complexity?}

\subsection{GNN-QE: Inductive Relational Structure Representation}
% find better motivation
% contribution: why GNN-QE is inductive
% zhaocheng: claim the original GNN-QE is a complex query model that encodes relative relational structure (can either be transductive or inductive). With proper parameterization, GNN-QE can be inductive, which aligns with our discussion of the essence of inductiveness.
% To the best of our knowledge, the only available query answering method that is inductive by design, can be trained on complex queries, and does not employ any entity embeddings is Graph Neural Network Query Executor (GNN-QE)~\cite{gnnqe}. 
% % how it's inductive
% Instead of learning explicit node representations in the encoder-decoder fashion, GNN-QE employs Neural Bellman-Ford network (NBFNet)~\cite{zhu2021neural} that captures relative relational structure uniquely for each simple link and complex query. \mg{How we achieve inductiveness}

The second strategy relies on learning inductive relational structure representations instead of explicit node representations. 
%Having the invariant set of relation types $\gR$ at training and inference time, relation embeddings are learned, and entities are parameterized through a relative relational structure of a given query conditioned on its starting anchor nodes. 
Having the same set of relation types $\gR$ at training and inference time, we can parameterize each entity based on the relative relational structure between it and the anchor nodes in a given query.
For instance (Fig.~\ref{fig:method} right), given a query with a particular relational structure \inlinegraphics{img/Y_arrows.pdf} and a set of anchor nodes, 
%all other nodes capture the relational structure relative to the anchor nodes. 
the representation of each node captures its relational structure relative to the anchor nodes.
Each neighborhood expansion step is equivalent to a \emph{projection} step in a complex query. In our example, immediate neighboring nodes will capture the intersection pattern \inlinegraphics{img/V_arrows.pdf}, and further nodes, in turn, capture the extended \emph{intersection-projection} structure \inlinegraphics{img/Y_arrows.pdf}.

Therefore, a node is likely to be an answer if its captured (or predicted) relational structure conforms with the query relational structure. 
As long as the set of relations is fixed, relation \emph{projection} is performed in the same way for training or new unseen nodes.
The idea of a one-hop (\emph{1p}) projection for simple link prediction has been proposed by Neural Bellman-Ford Networks (NBFNet)~\cite{zhu2021neural}. 

In particular, given a relation \emph{projection} query $(h, r, ?)$, NBFNet assigns unique initial states $\vh^{(0)}$ to all nodes in a graph by applying an indicator function $\vh^{(0)} = \textsc{Indicator}(h, v, r)$, i.e., a head node $h$ is initialized with a learnable relation embedding $\vr$ and all other nodes are initialized with zeros.
Then, NBFNet applies $L$ relational message passing GNN layers where each layer $l$ has its own learnable relation embedding matrix $\mR^{(l)} $ obtained as a projection (and reshaping) of thein initial relation $\mR^{(l)} = \mW^{(l)} \vr + \vb^{(l)} $. 
Final layer representations $\vh^{(L)}$ are passed through an MLP and the sigmoid function $\sigma$ to get a probability distribution over all nodes in a graph $p(t|h, r) = \sigma(\text{MLP}(\vh^{(L)}))$.
As each projection query spawns a uniquely initialized graph and message passing procedure, NBFNet is seen to be applying a \emph{labeling trick}~\cite{label_trick} to model a conditional probability distribution $p(t |h, r)$ that is provably more expressive than a joint distribution $p(h,r,t)$ produced by standard graph encoders.

% how NBFNet-QE processes logical ops
Applied to complex queries, chaining $k$ NBFNet instances allows us to answer $k$-hop projection queries, e.g., two instances for \emph{2p} queries. 
GNN-QE employs NBFNet as a \emph{trainable} projection operator and endows it with differentiable, non-parametric \emph{product} logic for modeling conjunction ($\land$), disjunction ($\lor$), and negation ($\lnot$) over the \emph{fuzzy sets} of all entities $\vx \in [0,1]^\gE$, i.e., after applying a logical operator (discussed in Appendix~\ref{app:tnorms}), each entity's degree of truth is associated with a scalar in range $[0, 1]$.
For the $i$-th hop projection, the indicator function initializes a node state $\vh_e^{(0)}$ with a relation vector $\vr_i$ weighted by a scalar probability predicted in the previous hop $x_e$: $\vh_e^{(0)} = x_e \vr_i$.
% what is trainable and what's at inference
%Differentiable projection and logical operators allow NBFNet-QE to be trained end-to-end on complex queries. 
Differentiable logical operators allow training GNN-QE end-to-end on complex queries. 
%\mg{Remove the last?} As each query is executed over a uniquely initialized instance of a graph, NBFNet-QE achieves the conditional answer parameterization $f(e|\gC, \gR_\gQ, \gG)$.

% GNN-QE - instantiating a graph with unique labeling pertaining to a triple / query building a conditional distribution $p(t | h, r)$, higher complexity, better expressiveness

% % THEORY - NOT NEEDED RIGHT NOW
% Possible theoretical basis: 
% \begin{itemize}
%     \item Logical queries we deal with have intersection ($\land$), union ($\lor$), negation ($\lnot$) operators and existentially quantified ($\exists$) variables. This is a subset of $\mathcal{ALCQ}$ description logic.
%     \item from \cite{Barcelo2020The} we know that \emph{homogeneous} Aggregate-Combine GNNs can represent any $\mathcal{ALCQ}$ logical formula, and Aggregate-Combine-Readout GNNs (with the final readout layer as in NBFNet and GNN-QE) can represent any $\text{FOC}_2$ formula
%     \item If we prove that \emph{heterogeneous} relational GNNs such as GNN-QE also fall within that theory, then we prove that GNN-QE can capture all logical queries (considered in our dataset) on the \emph{complete} graph without missing edges. Our experiments (Fig.\ref{fig:train_queries}) already confirm that training query accuracy is pretty much 1.0.
% \end{itemize}

\section{Experiments}

We designed the experimental agenda to demonstrate that inductive representation strategies are able to:
1) answer complex logical queries over new, unseen entities at inference time, i.e., when query anchors are new nodes (Section~\ref{sec:test_queries}); 
2) predict new correct answers for known \emph{training} queries when executed over larger inference graphs, i.e., when query anchors come from the training graph but variables and answers belong to the larger inference graph (Section~\ref{sec:train_queries});
3) generalize to inference graphs of up to 500\% larger than training graphs;
4) scale to inductive query answering over graphs of millions of nodes when updated with 500k new nodes and 5M new edges (Section~\ref{sec:scaling}).
%aiming at answering four research questions. \mg{Or re-frame that we can do the following}
% \textbf{RQ 1)} Can inductive representation learning methods perform complex logical reasoning over new, unseen entities at inference time? 
% \textbf{RQ 2)} Is training on complex query necessary or can inference-only methods yield non-trivial performance?
% \textbf{RQ 3)} Do inductive models scale to larger inference graphs?
% \textbf{RQ 4)} Can inductive models capture the extended answer set of training queries on bigger inference graphs?
% \mg{Can also add smth about theoretical expressiveness of NBFNet-QE.}

\subsection{Setup \& Dataset}
\label{sec:dataset}

\textbf{Dataset.} 
Due to the absence of inductive logical query benchmarks, we create a novel suite of datasets based on FB15k-237~\cite{toutanova-chen-2015-observed} (open license) and following the query generation process of BetaE~\cite{ren2020beta}.
Given a source graph with $\gE$ entities, we sample $|\gE_{\textit{train}}| = r \cdot |\gE|, r \in [0.1, 0.9]$ nodes to induce a training graph $\gG_{\textit{train}}$. 
For validation and test graphs, we split the remaining set of entities into two non-overlapping sets each with $\frac{1-r}{2}|\gE|$ nodes. 
We then merge training and unseen nodes into the inference set of nodes $\gE_{\textit{inf}}$ and induce inference graphs for validation and test from those sets, respectively, i.e., $\gE_{\textit{inf}}^{\textit{val}} = \gE_{\textit{train}} \cup \gE_{\textit{val}}$ and $\gE_{\textit{inf}}^{\textit{test}} = \gE_{\textit{train}} \cup \gE_{\textit{test}}$.
That is, validation and test inference graphs both extend the training graph but their sets of new entities are disjoint. Finally, we sample and remove 15\% of edges $\gT_{\text{pred}}$ in the inference graphs as missing edges for sampling queries with those missing edges.
Overall, we sample 9 such datasets based on different choices of $r$, which result in the ratios of inference graph size to the training graph $\gE_{\textit{inf}} / \gE_{\textit{train}}$ from 105\% to 550\%.

For each dataset, we employ the query sampler from BetaE~\cite{ren2020beta} to extract 14 typical query types \emph{1p/2p/3p/2i/3i/ip/pi/2u/up/2in/3in/inp/pin/pni}. 
Training queries are sampled from the training graph $\gG_{\textit{train}}$, validation and test queries are sampled from their respective inference graphs $\gG_{\textit{inf}}$ where at least one edge belongs to $\gT_{\text{pred}}$ and has to be predicted at inference time.

As inference graphs extend training graphs, training queries are very likely to have new answers when executed over $\gG_{\textit{inf}}$ with simple graph traversal and without any link prediction. We create an additional set of true answers for all training queries executed over the test inference graph $\gG_{\textit{inf}}^{\textit{test}}$ to measure the entailment capabilities of query answering models. 
This is designed to be an inference task and extends the \emph{faithfullness} evaluation of \cite{DBLP:conf/nips/SunAB0C20}.
Dataset statistics can be found in Appendix~\ref{app:dataset}.

\textbf{Evaluation Protocol.}
Following the literature~\cite{ren2020beta}, query answers are separated into two sets: \emph{easy answers} that only require graph traversal over existing edges, and \emph{hard answers} that require inferring missing links to achieve the answer node. 
For the main experiment, evaluation involves ranking of \emph{hard} answers against all entities having easy ones filtered out. 
For evaluating training queries on inference graphs, we only have \emph{easy} answers and rank them against all entities. 
We report Hits@10 as the main performance metric on different query types.

\textbf{Implementation Details.}
All NodePiece-based models~\cite{galkin2022nodepiece} were pre-trained until convergence on a simple \emph{1p} link prediction task with the relations-only vocabulary and entity tokenization, MLP encoder, and ComplEx~\cite{trouillon2016complex} scoring function. 
We used a 2-layer CompGCN~\cite{Vashishth2020Composition-based} as an optional message passing encoder on top of NodePiece features. 
The non-parametric CQD-Beam~\cite{arakelyan2020complex} decoder for answering complex queries is tuned for each query type based on the validation set of queries, most of the setups employ a \emph{product t-norm}, sigmoid entity score normalization, and beam size of 32. 
Following the literature, the GNN-QE models~\cite{gnn_qe} were trained on 10 query patterns (\emph{1p/2p/3p/2i/3i/2in/3in/inp/pin/pni}) where \emph{ip/pi/2u/up} are only seen at inference time. 
Each model employs a 4-layer NBFNet~\cite{zhu2021neural} as a trainable projection operator with DistMult~\cite{DBLP:journals/corr/YangYHGD14} composition function and PNA~\cite{corso2020principal} aggregation. 
Other logical operators ($\land , \lor, \lnot$) are executed with the non-parametric \emph{product t-norm} and \emph{t-conorm}.
Both NodePiece-QE and GNN-QE are implemented\footnote{Code and data are available at \url{https://github.com/DeepGraphLearning/InductiveQE}} with PyTorch~\cite{DBLP:conf/nips/PaszkeGMLBCKLGA19} and trained with the Adam~\cite{DBLP:journals/corr/KingmaB14} optimizer. NodePiece-QE models were pre-trained and evaluated on a single Tesla V100 32 GB GPU whereas GNN-QE models were trained and evaluated on 4 Tesla V100 16GB.
All hyperparameters are listed in Appendix~\ref{app:hyperparams}.
To show that the proposed models are non-trivial, we compare them with an \emph{Edge-type Heuristic} baseline (Appendix~\ref{app:dummy}), which selects all entities that satisfy the relations in the last hop of the query in $\gG_{\textit{inf}}$.

% As a trivial baseline, we implement an \emph{Edge-type Heuristic} (Appendix~\ref{app:dummy}) that selects all tails sharing the incoming query relation $r$ of the projection query $(h, r, ?)$ from a \emph{relation-entity} dictionary precomputed on $\gG_{\textit{inf}}$.

\subsection{Complex Query Answering over Unseen Entities on Differently Sized Inference Graphs}
\label{sec:test_queries}

\begin{figure}[t]
    \centering
    \includegraphics[width=\textwidth]{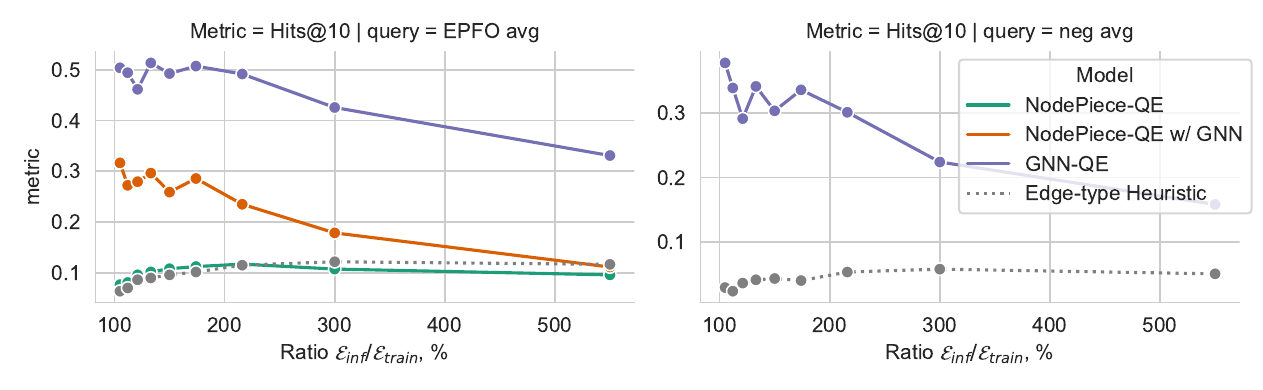}
    \caption{Aggregated Hits@10 performance of \textbf{test queries} (involving unseen entities) executed on inference graphs of different ratios compared to training graphs. NodePiece-based models are \emph{inference-only} and support EPFO queries, GNN-QE is trainable and supports negation queries. }
    \label{fig:main_exp}
\end{figure}

% Hits @ 10
\begin{table}[t]
    \centering
    \caption{Test Hits@10 results (\%) on answering inductive FOL queries when $\gE_{\textit{inf}} / \gE_{\textit{train}} = 175\%$. avg$_p$ is the average on EPFO queries ($\land$, $\lor$). avg$_n$ is the average on queries with negation.}
    \begin{adjustbox}{width=\textwidth}
    \begin{tabular}{lrrrrrrrrrrrrrrrr}
        \toprule
        \bf{Model} & \bf{avg$_p$} & \bf{avg$_n$} & \bf{1p} & \bf{2p} & \bf{3p} & \bf{2i} & \bf{3i} & \bf{pi} & \bf{ip} & \bf{2u} & \bf{up} & \bf{2in} & \bf{3in} & \bf{inp} & \bf{pin} & \bf{pni} \\
        \midrule
        \multicolumn{17}{c}{\emph{Transductive}} \\
        \midrule
        BetaE & 1.3 & 0.2 & 2.9 & 0.4 & 0.4 & 2.1 & 3.3 & 1.5 & 0.7 & 0.2 & 0.2 & 0.1 & 0.2 & 0.2 & 0.1 & 0.1 \\
        \midrule
        \multicolumn{17}{c}{\emph{Inductive Inference-only}} \\
        \midrule
        % NodePiece-QE & 11.0 & - & 21.3 & 8.9 & 5.1 & 13.0 & 14.7 & 9.8 & 8.7 & 9.7 & 7.5 & - & - & - & - & - \\
        % NodePiece-QE w/ GNN & 20.9 & - & 34.4 & 15.6 & 10.5 & 28.7 & 33.4 & 19.2 & 16.2 & 17.8 & 12.2 & - & - & - & - & - \\
        Edge-type Heuristic & 10.1 & 4.1 & 17.7 & 8.2 & 9.9 & 10.7 & 13.0 & 9.8 & 8.2 & 5.3 & 8.5 & 2.6 & 2.9 & 8.4 & 3.8 & 2.7 \\
        NodePiece-QE & 11.2 & - & 25.5 & 8.2 & 8.4 & 12.4 & 13.9 & 9.9 & 8.7 & 7.0 & 6.8 & - & - & - & - & - \\
        NodePiece-QE w/ GNN & 28.6 & - & 45.9 & 19.2 & 11.5 & 39.9 & 48.8 & 29.4 & 22.6 & 25.3 & 14.6 & - & - & - & - & - \\
        \midrule
        \multicolumn{17}{c}{\emph{Inductive Trainable}} \\
        \midrule
        GNN-QE & 50.7 & 33.6 & 65.4 & 36.3 & 31.6 & 73.8 & 84.3 & 56.5 & 41.5 & 39.3 & 28.0 & 33.3 & 46.4 & 29.2 & 24.9 & 34.0 \\
        \bottomrule
    \end{tabular}
    \end{adjustbox}
    \label{tab:main}
\end{table}

First, we probe \emph{inference-only} NodePiece-based embedding models and \emph{trainable}  GNN-QE in the inductive setup, i.e., query answering over unseen nodes requiring link prediction over unseen nodes. 
As a sanity check, we compare them to the Edge-type Heuristic and a transductive BetaE model~\cite{ren2020beta} 
%that is trained on the training graph and use randomly initialized embeddings for unseen nodes. 
 trained with standard hyperparameters (Appendix~\ref{app:hyperparams}) on the reference dataset (with ratio $\gE_{\textit{inf}} / \gE_{\textit{train}}$ of 175\%) with randomly initialized embeddings for unseen nodes at inference time.
Table~\ref{tab:main} summarizes the results on the reference dataset while Fig.~\ref{fig:main_exp} illustrates a bigger picture on all datasets (we provide a detailed breakdown by query type for all splits in Appendix~\ref{app:more_experiments}).
The experiment on the tranductive BetaE confirms that pure transductive models can not generalize to graphs with unseen nodes. 

With inductive models, however, we observe that even inference-only models pre-trained solely on simple \emph{1p} link prediction exhibit non-trivial performance in answering queries with unseen entities.
Particularly, the inference-only \emph{NodePiece with GNN} baseline exhibits better performance over all query types and inference graphs up to 300\% larger than training graphs.

The trainable GNN-QE models expectedly outperform non-trainable baselines and can tackle queries with negation ($\lnot$).
Here, we confirm that the \emph{labeling trick}~\cite{label_trick} and conditional $p(t|h,r)$ modeling better capture the relation projection problem than joint $p(h,r,t)$ encoding approaches.

Still, % all evaluated models with message passing,
we notice that models with GNNs, i.e., inference-only NodePiece-QE with GNN and trainable GNN-QE, suffer from increasing the size of the inference graph and having more unseen entities. 
Reaching best results on $\gE_{\textit{inf}} / \gE_{\textit{train}}$ ratios around 130\%, both approaches steadily deteriorate up until final 550\% by 20 absolute Hits@10 points on EPFO queries and negation queries. 
We attribute this deterioration to the known generalization issues~\cite{DBLP:conf/nips/KnyazevTA19, DBLP:conf/icml/YehudaiFMCM21} of message passing GNNs when performing inference over a larger graph than the network has seen during training. 
Recently, a few strategies have been proposed~\cite{buffelli2022sizeshiftreg,zhou2022ood} to alleviate this issue and we see it as a promising avenue for future work.
On the other hand, a simple NodePiece-QE model without message passing retains similar performance independently of the inference graph size.

Lastly, we  observe that lower performance of inference-only NodePiece models can be also attributed to underfitting (cf.~train graph charts in Fig.~\ref{fig:train_queries}). 
Although \emph{1p} link predictors were trained until convergence (on the inductive validation set of missing triples), the performance of training queries on training graphs with \emph{easy answers} that require only relation traversal without predicting missing edges is not yet saturated. 
This fact suggests that better fitting entity featurization (obtained by NodePiece or other strategies) could further improve the test performance in the inference-only regime. 
We leave the search of such strategies for future work.

% Hits @ 3
% \begin{table}[t]
%     \centering
%     \caption{Test Hits@3 results (\%) on answering FOL queries when $\gE_{\textit{inf}} / \gE_{\textit{train}} = 175\%$. avg$_p$ is the average Hits@3 on EPFO queries ($\land$, $\lor$). avg$_n$ is the average Hits@3 on queries with negation.}
%     \begin{adjustbox}{width=\textwidth}
%     \begin{tabular}{lcccccccccccccccc}
%         \toprule
%         \bf{Model} & \bf{avg$_p$} & \bf{avg$_n$} & \bf{1p} & \bf{2p} & \bf{3p} & \bf{2i} & \bf{3i} & \bf{pi} & \bf{ip} & \bf{2u} & \bf{up} & \bf{2in} & \bf{3in} & \bf{inp} & \bf{pin} & \bf{pni} \\
%         \midrule
%         \multicolumn{17}{c}{\emph{Inference-only}} \\
%         \midrule
%         NodePiece + CQD & 4.8 & - & 9.8 & 4.1 & 2.1 & 5.5 & 6.0 & 3.9 & 3.9 & 4.4 & 3.3 & - & - & - & - & - \\
%         NodePiece + GNN + CQD & 11.2 & - & 19.5 & 7.5 & 4.7 & 15.3 & 20.7 & 9.4 & 7.9 & 9.8 & 5.7 & - & - & - & - & - \\
%         \midrule
%         \multicolumn{17}{c}{\emph{Trainable}} \\
%         \midrule
%         GNN-QE & 38.9 & 4.8 & 54.3 & 27.9 & 20.5 & 59.1 & 72.2 & 44.7 & 29.7 & 24.4 & 17.8 & 4.5 & 7.7 & 8.9 & 1.5 & 1.5 \\
%         \bottomrule
%     \end{tabular}
%     \end{adjustbox}
%     \label{tab:main}
% \end{table}

\subsection{Predicting New Answers for Training Queries on Larger Inference Graphs}
\label{sec:train_queries}

Simulating the incremental addition of new edges in graph databases, we evaluate the performance of our inference-only and trainable QE models on \emph{training} queries on the original training graph and extended inference graph (with added test edges).
As databases are able to immediately retrieve new answers to known queries after updating the graph, we aim at exploring and quantifying this behaviour of neural reasoning models.
In this experiment, we probe training queries and their \emph{easy answers} that require performing only graph traversal without predicting missing links in the inference graph. 
While execution of training queries over the \emph{training} graph indicates how well the model could fit training data, executing training queries over the bigger \emph{inference} graph with new entities aims to capture basic reasoning capabilities of QE models in the inductive regime.

Particular challenges arising when executing training queries over a bigger graph are: (1) the same queries can have more correct answers as more new nodes and edges satisfying the query pattern might have been added (as in Fig.~\ref{fig:intro1}); (2) more new entities create a ``distractor'' setting with more false positives. 
Generally, evaluation of training queries on the inference graph can be considered as an extended version of the \emph{faithfullness}~\cite{DBLP:conf/nips/SunAB0C20} evaluation that captures how well a trained model can answer original training queries, i.e., memorization capacity.
In all 9 datasets, most of training queries have at least one new correct answer in the inference graph (more details in Appendix~\ref{app:dataset}).

\begin{figure}[t]
    \centering
    \includegraphics[width=\textwidth]{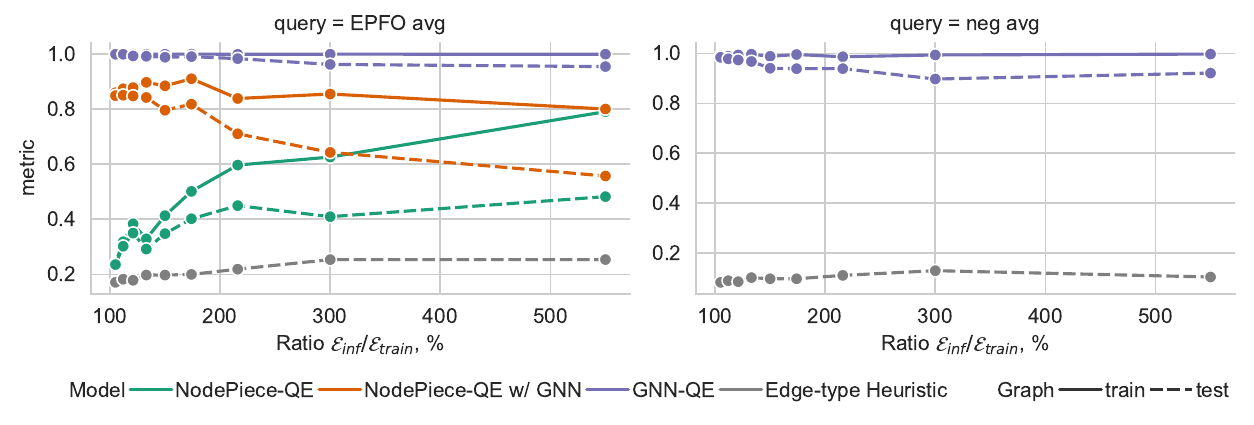}
    \caption{Aggregated Hits@10 performance of \textbf{training queries} on the original training and extended test inference graphs where queries have new correct answers. NodePiece-based models are \emph{inference-only} and support EPFO queries, GNN-QE is trainable and supports negation queries.}
    \label{fig:train_queries}
\end{figure}

Fig.~\ref{fig:train_queries} illustrates the performance of the Edge-type Heuristic baseline, inference-only NodePiece-QE (without and with GNN) and trainable GNN-QE models.
Generally, GNN-QE fits the training query data almost perfectly confirming the original finding~\cite{zhu2021neural} that NBFNet performs graph traversal akin to symbolic models. 
GNN-QE can also find new correct answers on graphs up to $300\%$ larger than training ones.
Then, the performance deteriorates which we attribute to the \emph{distractor} factor with more unseen entities and the mentioned generalization issue on larger inference graphs.% and generalization issues~\cite{DBLP:conf/nips/KnyazevTA19, DBLP:conf/icml/YehudaiFMCM21} of message passing GNNs when performing inference over much larger graph than the network has seen during training.

The inference-only NodePiece-QE models, as expected, do not fully fit the training data as they were never trained on complex queries. 
Still, the inference-only models exhibit non-trivial performance (compared to the Edge-type Heuristic) in finding more answers on graphs up to $200\%$ larger than training ones with relatively small performance margins compared to training queries. 
%This fact supports the hypothesis in Sec.~\ref{sec:test_queries} of improving the performance with better entity featurization strategies.
The most surprising observation is that GNN-free NodePiece-QE models improve the performance on both training and inference graphs as the graphs (and the $\gE_{\textit{inf}} / \gE_{\textit{train}}$ ratio) grow larger while GNN-enriched models steadily deteriorate. 
We attribute this growth to the relation-based NodePiece tokenization and its learned features that tend to be more discriminative in larger inference graphs where new nodes have smaller degree and thus can be better identified by their incident relation types.
We provide more experimental results for each dataset ratio with breakdown by query type  in Appendix~\ref{app:more_experiments}.

\subsection{Ranking of Easy and Hard Answers}
\label{sec:easy_vs_hard}

In addition to evaluating \emph{faithfullness} that measures whether a model could recover easy answers, it is also insightful to measure whether all easy answers can be ranked higher than hard answers. That is, a reliable query answering model would first recover all possible easy answers and would enrich the answer set with highly-probable hard answers.
To this end, we apply a ROC AUC metric over original \textbf{unfiltered} scores.
% The idea of computing ROC AUC is as follows:
% having a list of unfiltered raw scores, we extract scores of all easy and hard answers. Let a query have 4 easy and 1 hard answer: $[5, 6, 7, 8, 32]$  where 8 is a rank of a hard answer while 5, 6, 7, 32 are ranks of easy answers. 
% We then create binary labels for the scores assigning 1 to the hard answers, e.g., $[0, 0, 0, 1, 0]$.
%Given those two arrays, we then compute 
The ROC AUC score measures how many hard answers are ranked \emph{after} easy answers.
%, e.g., in our example ROC AUC is 0.75. 
Note that the score does not depend on actual values of ranks, that is, the metric will be high when easy answers are ranked, e.g., in between 100-1000 as long as hard answers are ranked 1001 and lower. 
Therefore, ROC AUC still needs to be paired with MRR to see how easy and hard answers are ranked absolutely.

\begin{table}[!ht]
    \centering
    \caption{Macro-averaged ROC AUC score over \textbf{unfiltered} predictions on the reference $\gE_{\textit{inf}} / \gE_{\textit{train}} = 175\%$ dataset to measure if all easy answers are ranked higher than hard answers. Higher is better.}
    \begin{adjustbox}{width=\textwidth}
    \begin{tabular}{lcccccccccccccccc}
        \toprule
        \bf{Model} & \bf{avg$_p$} & \bf{avg$_n$} & \bf{1p} & \bf{2p} & \bf{3p} & \bf{2i} & \bf{3i} & \bf{pi} & \bf{ip} & \bf{2u} & \bf{up} & \bf{2in} & \bf{3in} & \bf{inp} & \bf{pin} & \bf{pni} \\ \midrule
        \multicolumn{17}{c}{\emph{Inductive Inference-only}} \\
        \midrule
        NodePiece-QE & 0.692 & & 0.623 & 0.710 & 0.711 & 0.657 & 0.654 & 0.692 & 0.731 & 0.723 & 0.729 & & & & & \\
NodePiece-QE w/ GNN & 0.776 & & 0.783 & 0.783 & 0.739 & 0.758 & 0.733 & 0.760 & 0.801 & 0.841 & 0.787 & & & & & \\
        \midrule
        \multicolumn{17}{c}{\emph{Inductive Trainable}} \\
        \midrule
        GNN-QE & 0.973 & 0.885 & 0.998 & 0.992 & 0.986 & 0.969 & 0.962 & 0.967 & 0.969 & 0.938 & 0.978 & 0.879 & 0.859 & 0.926 & 0.914 & 0.847 \\
        \bottomrule
    \end{tabular}
    \end{adjustbox}
    \label{tab:rocauc}
\end{table}

We compute ROC AUC for each query and average them over each query type thus making it \textbf{macro-averaged ROC AUC}. 
Our experimental results on all query types using the models reported in Table~\ref{tab:main} on the reference 175\% dataset are compiled in Table~\ref{tab:rocauc}.

GNN-QE has nearly perfect ROC AUC scores as it was trained on complex queries. 
NodePiece-QE models are acceptable for inference-only models that were only trained only on 1p simple link prediction and have never seen any complex query at training time.

\subsection{Scaling to Millions of Nodes on WikiKG-QE}
\label{sec:scaling}

Finally, we perform a scalability experiment evaluating complex query answering in the inductive mode on a new large dataset \emph{WikiKG-QE} constructed from OGB WikiKG 2~\cite{DBLP:conf/nips/HuFZDRLCL20} (CC0 license).
While the original task is transductive link prediction, we split the graph into a training graph of 1.5M entities (5.8M edges, 512 unique relation types) and  validation (test) graphs of 500k unseen nodes (5M known and 600k missing edges) each. 
The resulting validation (test) inference graphs are therefore of 2M entities and 11M edges with the $\gE_{\textit{inf}} / \gE_{\textit{train}}$ ratio of $133\%$ (details are in Appendix~\ref{app:dataset}). 

GNN-QE cannot scale to such sizes, so we only probe NodePiece-QE models.
Due to the problem size, we only sample 10k EPFO queries of each type from the \emph{test inference} graph to run in the inference-only regime. Each query has at least one missing edge to be predicted at inference.
The answers are ranked against all 2M entities in the filtered setting (in contrast to the OGB task that ranks against 1000 pre-computed negative samples) and Hits@100 as the target metric.

We pre-train a NodePiece encoder (in addition to relation types, we tokenize nodes with a vocabulary of 20k nodes, total 3M parameters in the encoder) with the ComplEx decoder on \emph{1p} link prediction over the training graph for 1M steps (see Appendix~\ref{app:hyperparams} for hyperparameters).
Then, the graph is extended with 500k new nodes and 5M new edges forming the inference graph.
Then, using the pre-trained encoder, we materialize representations of entities (both seen and new) and relations from this inference graph.
Finally, CQD-Beam executes the queries against the bigger inference graph extended with 500k new nodes and 5M new edges.

\begin{table}[!h]
    \centering
    \caption{Test Hits@100 of NodePiece-QE on WikiKG-QE (2M nodes, 11M edges including 500k new nodes and 5M new edges) in the inference-only regime. avg$_p$ is the average on EPFO queries.}
    %\begin{adjustbox}{width=\textwidth}
    \begin{tabular}{lcccccccccccccccc}
        \toprule
        \bf{Model} & \bf{avg$_p$} & \bf{1p} & \bf{2p} & \bf{3p} & \bf{2i} & \bf{3i} & \bf{pi} & \bf{ip} & \bf{2u} & \bf{up} \\
        \midrule
        %\textcolor{blue}{OLD: NodePiece + CQD} & 3.1 & 12.4 & 1.2 & 0.8 & 2.4 & 4.6 & 1.7 & 1.0 & 2.8 & 0.7 \\
        %NodePiece-QE & 6.6 & 19.0 & 2.1 & 1.8 & 11.0 & 15.8 & 4.4 & 2.8 & 1.3 & 1.5 \\
        Edge-type Heuristic & 3.1 & 10.0 & 1.0 & 0.9 & 3.7 & 8.1 & 1.8 & 0.9 & 0.7 & 0.5 \\
        NodePiece-QE & 9.2 & 22.6 & 5.2 & 3.9 & 11.6 & 17.4 & 7.0 & 4.5 & 7.4 & 3.2 \\
        NodePiece-QE w/ GNN & 10.1 & 66.6 & 0.9 & 0.6 & 5.4 & 8.2 & 2.3 & 0.8 & 5.2 & 0.5 \\
        \bottomrule
    \end{tabular}
    %\end{adjustbox}
    \label{tab:wikikg}
\end{table}

As shown in Table~\ref{tab:wikikg}, we find a non-trivial performance of the inference-only model on EPFO queries demonstrating that inductive \emph{node representation} QE models are able to scale to graphs with hundreds of thousands of new nodes and millions of new edges in the zero-shot fashion.
That is, answering complex queries over unseen entities is available right upon updating the graph without the need to retrain a model.
This fact paves the way for the concept of \emph{neural graph databases} capable of performing zero-shot inference over updatable graphs without expensive retraining.  
 
\section{Limitations and Future Work}
\label{sec:conclusion}

\textbf{Limitations.}
With the two proposed inductive query answering strategies, we observe a common trade-off between the performance and computational complexity. That is, inductive \emph{node representation} models like NodePiece-QE are fast, scalable, and can be executed in the inference-only regime but underperform compared to the inductive \emph{relational structure representation} models like GNN-QE. On the other hand, GNN-QE incurs high computational costs due to executing each query on a uniquely initialized graph instance. Alleviating this issue is a key to scalability.

\textbf{Societal Impact.}
The inductive setup assumes running inference on (partly) unseen data, that is, the nature of this unseen data might be out-of-distrbution, unknown and potentially malicious. This fact has to be taken into account when evaluating predictions and overall system trustworthiness.

\textbf{Conclusion and Future Work.}
In this work, we defined the problem of inductive complex logical query answering and proposed two possible parameterization strategies based on \emph{node} and \emph{relational structure} representations to deal with new, unseen entities at inference time. 
Experiments demonstrated that both strategies are able to answer complex logical queries over unseen entities as well as identify new answers on larger inference graphs. 
In the future work, we plan to extend the inductive setup to completely disjoint training and inference graphs, expand the set of supported logical query patterns aligned with popular queries over real-world KGs, enable reasoning over continuous features like texts and numbers, support more KG modalities like hypergraphs and hyper-relational graphs, and further explore the concept of neural graph databases.

\begin{ack}
This project is supported by the Natural Sciences and Engineering Research Council (NSERC) Discovery Grant, the Canada CIFAR AI Chair Program, collaboration grants between Microsoft Research and Mila, Samsung Electronics Co., Ltd., Amazon Faculty Research Award, Tencent AI Lab Rhino-Bird Gift Fund and a NRC Collaborative R\&D Project (AI4D-CORE-06). This project was also partially funded by IVADO Fundamental Research Project grant PRF-2019-3583139727. The computation resource of this project is supported by Calcul Qu\'ebec\footnote{\url{https://www.calculquebec.ca/}} and Compute Canada\footnote{\url{https://www.computecanada.ca/}}.
We would like to thank anonymous reviewers for the comments that helped to improve the manuscript.
\end{ack}

\bibliography{bibliography}

\begin{thebibliography}{10}

\bibitem{ali2020light}
Mehdi Ali, Max Berrendorf, Charles~Tapley Hoyt, Laurent Vermue, Mikhail Galkin,
  Sahand Sharifzadeh, Asja Fischer, Volker Tresp, and Jens Lehmann.
\newblock Bringing light into the dark: {A} large-scale evaluation of knowledge
  graph embedding models under a unified framework.
\newblock {\em IEEE Transactions on Pattern Analysis and Machine Intelligence},
  2021.

\bibitem{ali2021pykeen}
Mehdi Ali, Max Berrendorf, Charles~Tapley Hoyt, Laurent Vermue, Sahand
  Sharifzadeh, Volker Tresp, and Jens Lehmann.
\newblock {PyKEEN 1.0: A Python Library for Training and Evaluating Knowledge
  Graph Embeddings}.
\newblock {\em Journal of Machine Learning Research}, 22(82):1--6, 2021.

\bibitem{alivanistos2022query}
Dimitrios Alivanistos, Max Berrendorf, Michael Cochez, and Mikhail Galkin.
\newblock Query embedding on hyper-relational knowledge graphs.
\newblock In {\em International Conference on Learning Representations}, 2022.

\bibitem{amayuelas2022neural}
Alfonso Amayuelas, Shuai Zhang, Xi~Susie Rao, and Ce~Zhang.
\newblock Neural methods for logical reasoning over knowledge graphs.
\newblock In {\em International Conference on Learning Representations}, 2022.

\bibitem{arakelyan2020complex}
Erik Arakelyan, Daniel Daza, Pasquale Minervini, and Michael Cochez.
\newblock Complex query answering with neural link predictors.
\newblock In {\em International Conference on Learning Representations}, 2021.

\bibitem{DBLP:conf/nips/BordesUGWY13}
Antoine Bordes, Nicolas Usunier, Alberto Garc{\'{\i}}a{-}Dur{\'{a}}n, Jason
  Weston, and Oksana Yakhnenko.
\newblock Translating embeddings for modeling multi-relational data.
\newblock In {\em Advances in Neural Information Processing Systems 26: 27th
  Annual Conference on Neural Information Processing Systems 2013}, pages
  2787--2795, 2013.

\bibitem{buffelli2022sizeshiftreg}
Davide Buffelli, Pietro Li{\`o}, and Fabio Vandin.
\newblock {SizeShiftReg}: a regularization method for improving
  size-generalization in graph neural networks.
\newblock In {\em Advances in Neural Information Processing Systems}, 2022.

\bibitem{chen2021fuzzy}
Xuelu Chen, Ziniu Hu, and Yizhou Sun.
\newblock Fuzzy logic based logical query answering on knowledge graph.
\newblock In {\em International Conference on Machine Learning}. PMLR, 2021.

\bibitem{choudhary2021probabilistic}
Nurendra Choudhary, Nikhil Rao, Sumeet Katariya, Karthik Subbian, and
  Chandan~K. Reddy.
\newblock Probabilistic entity representation model for reasoning over
  knowledge graphs.
\newblock In {\em Thirty-Fifth Conference on Neural Information Processing
  Systems}, 2021.

\bibitem{corso2020principal}
Gabriele Corso, Luca Cavalleri, Dominique Beaini, Pietro Li{\`o}, and Petar
  Veli{\v{c}}kovi{\'c}.
\newblock Principal neighbourhood aggregation for graph nets.
\newblock {\em Advances in Neural Information Processing Systems},
  33:13260--13271, 2020.

\bibitem{daza2020message}
Daniel Daza and Michael Cochez.
\newblock Message passing query embedding.
\newblock {\em arXiv preprint arXiv:2002.02406}, 2020.

\bibitem{Fey/Lenssen/2019}
Matthias Fey and Jan~E. Lenssen.
\newblock Fast graph representation learning with {PyTorch Geometric}.
\newblock In {\em ICLR Workshop on Representation Learning on Graphs and
  Manifolds}, 2019.

\bibitem{galkin2022nodepiece}
Mikhail Galkin, Etienne Denis, Jiapeng Wu, and William~L. Hamilton.
\newblock Nodepiece: Compositional and parameter-efficient representations of
  large knowledge graphs.
\newblock In {\em International Conference on Learning Representations}, 2022.

\bibitem{DBLP:conf/icml/GilmerSRVD17}
Justin Gilmer, Samuel~S. Schoenholz, Patrick~F. Riley, Oriol Vinyals, and
  George~E. Dahl.
\newblock Neural message passing for quantum chemistry.
\newblock In {\em Proceedings of the 34th International Conference on Machine
  Learning, {ICML} 2017}, volume~70 of {\em Proceedings of Machine Learning
  Research}, pages 1263--1272. {PMLR}, 2017.

\bibitem{hamilton2018embedding}
Will Hamilton, Payal Bajaj, Marinka Zitnik, Dan Jurafsky, and Jure Leskovec.
\newblock Embedding logical queries on knowledge graphs.
\newblock {\em Advances in Neural Information Processing Systems}, 31, 2018.

\bibitem{DBLP:conf/nips/HuFZDRLCL20}
Weihua Hu, Matthias Fey, Marinka Zitnik, Yuxiao Dong, Hongyu Ren, Bowen Liu,
  Michele Catasta, and Jure Leskovec.
\newblock Open graph benchmark: Datasets for machine learning on graphs.
\newblock In {\em Advances in Neural Information Processing Systems 33: Annual
  Conference on Neural Information Processing Systems 2020}, 2020.

\bibitem{ji2022kgs}
Shaoxiong Ji, Shirui Pan, Erik Cambria, Pekka Marttinen, and Philip~S. Yu.
\newblock A survey on knowledge graphs: Representation, acquisition and
  applications.
\newblock {\em IEEE Transactions on Neural Networks and Learning Systems},
  33(2):494--514, 2022.

\bibitem{DBLP:journals/corr/KingmaB14}
Diederik~P. Kingma and Jimmy Ba.
\newblock Adam: A method for stochastic optimization.
\newblock In {\em ICLR (Poster)}, 2015.

\bibitem{DBLP:journals/fss/KlementMP04a}
Erich{-}Peter Klement, Radko Mesiar, and Endre Pap.
\newblock Triangular norms. position paper {I:} basic analytical and algebraic
  properties.
\newblock {\em Fuzzy Sets Syst.}, 143(1):5--26, 2004.

\bibitem{DBLP:conf/nips/KnyazevTA19}
Boris Knyazev, Graham~W. Taylor, and Mohamed~R. Amer.
\newblock Understanding attention and generalization in graph neural networks.
\newblock In {\em Advances in Neural Information Processing Systems 32: Annual
  Conference on Neural Information Processing Systems 2019, NeurIPS 2019},
  pages 4204--4214, 2019.

\bibitem{biqe}
Bhushan Kotnis, Carolin Lawrence, and Mathias Niepert.
\newblock Answering complex queries in knowledge graphs with bidirectional
  sequence encoders.
\newblock {\em CoRR}, abs/2004.02596, 2020.

\bibitem{DBLP:conf/nips/PaszkeGMLBCKLGA19}
Adam Paszke, Sam Gross, Francisco Massa, Adam Lerer, James Bradbury, Gregory
  Chanan, Trevor Killeen, Zeming Lin, Natalia Gimelshein, Luca Antiga, Alban
  Desmaison, Andreas K{\"{o}}pf, Edward Yang, Zachary DeVito, Martin Raison,
  Alykhan Tejani, Sasank Chilamkurthy, Benoit Steiner, Lu~Fang, Junjie Bai, and
  Soumith Chintala.
\newblock Pytorch: An imperative style, high-performance deep learning library.
\newblock In {\em Advances in Neural Information Processing Systems 32: Annual
  Conference on Neural Information Processing Systems 2019, NeurIPS 2019},
  pages 8024--8035, 2019.

\bibitem{ren2019query2box}
Hongyu Ren, Weihua Hu, and Jure Leskovec.
\newblock Query2box: Reasoning over knowledge graphs in vector space using box
  embeddings.
\newblock In {\em International Conference on Learning Representations}, 2019.

\bibitem{ren2020beta}
Hongyu Ren and Jure Leskovec.
\newblock Beta embeddings for multi-hop logical reasoning in knowledge graphs.
\newblock {\em Advances in Neural Information Processing Systems}, 33, 2020.

\bibitem{sadeghian2019drum}
Ali Sadeghian, Mohammadreza Armandpour, Patrick Ding, and Daisy~Zhe Wang.
\newblock Drum: End-to-end differentiable rule mining on knowledge graphs.
\newblock {\em Advances in Neural Information Processing Systems}, 32, 2019.

\bibitem{DBLP:conf/nips/SunAB0C20}
Haitian Sun, Andrew~O. Arnold, Tania Bedrax{-}Weiss, Fernando Pereira, and
  William~W. Cohen.
\newblock Faithful embeddings for knowledge base queries.
\newblock In {\em Advances in Neural Information Processing Systems 33: Annual
  Conference on Neural Information Processing Systems 2020, NeurIPS 2020},
  2020.

\bibitem{DBLP:conf/iclr/SunDNT19}
Zhiqing Sun, Zhi{-}Hong Deng, Jian{-}Yun Nie, and Jian Tang.
\newblock Rotate: Knowledge graph embedding by relational rotation in complex
  space.
\newblock In {\em 7th International Conference on Learning Representations,
  {ICLR} 2019}. OpenReview.net, 2019.

\bibitem{DBLP:conf/icml/TeruDH20}
Komal~K. Teru, Etienne Denis, and Will Hamilton.
\newblock Inductive relation prediction by subgraph reasoning.
\newblock In {\em Proceedings of the 37th International Conference on Machine
  Learning, {ICML} 2020}, volume 119 of {\em Proceedings of Machine Learning
  Research}, pages 9448--9457. {PMLR}, 2020.

\bibitem{toutanova-chen-2015-observed}
Kristina Toutanova and Danqi Chen.
\newblock Observed versus latent features for knowledge base and text
  inference.
\newblock In {\em Proceedings of the 3rd Workshop on Continuous Vector Space
  Models and their Compositionality}, pages 57--66, Beijing, China, July 2015.
  Association for Computational Linguistics.

\bibitem{trouillon2016complex}
Th{\'e}o Trouillon, Johannes Welbl, Sebastian Riedel, {\'E}ric Gaussier, and
  Guillaume Bouchard.
\newblock Complex embeddings for simple link prediction.
\newblock In {\em International conference on machine learning}, pages
  2071--2080. PMLR, 2016.

\bibitem{Vashishth2020Composition-based}
Shikhar Vashishth, Soumya Sanyal, Vikram Nitin, and Partha Talukdar.
\newblock Composition-based multi-relational graph convolutional networks.
\newblock In {\em International Conference on Learning Representations}, 2020.

\bibitem{DBLP:conf/www/WestGMSGL14}
Robert West, Evgeniy Gabrilovich, Kevin Murphy, Shaohua Sun, Rahul Gupta, and
  Dekang Lin.
\newblock Knowledge base completion via search-based question answering.
\newblock In {\em 23rd International World Wide Web Conference, {WWW} '14},
  pages 515--526. {ACM}, 2014.

\bibitem{DBLP:journals/corr/YangYHGD14}
Bishan Yang, Wen{-}tau Yih, Xiaodong He, Jianfeng Gao, and Li~Deng.
\newblock Embedding entities and relations for learning and inference in
  knowledge bases.
\newblock In {\em 3rd International Conference on Learning Representations,
  {ICLR} 2015}, 2015.

\bibitem{DBLP:conf/nips/YangYC17}
Fan Yang, Zhilin Yang, and William~W. Cohen.
\newblock Differentiable learning of logical rules for knowledge base
  reasoning.
\newblock In {\em Advances in Neural Information Processing Systems 30: Annual
  Conference on Neural Information Processing Systems 2017}, pages 2319--2328,
  2017.

\bibitem{DBLP:conf/icml/YehudaiFMCM21}
Gilad Yehudai, Ethan Fetaya, Eli~A. Meirom, Gal Chechik, and Haggai Maron.
\newblock From local structures to size generalization in graph neural
  networks.
\newblock In {\em Proceedings of the 38th International Conference on Machine
  Learning, {ICML} 2021}, volume 139 of {\em Proceedings of Machine Learning
  Research}, pages 11975--11986. {PMLR}, 2021.

\bibitem{label_trick}
Muhan Zhang, Pan Li, Yinglong Xia, Kai Wang, and Long Jin.
\newblock Labeling trick: A theory of using graph neural networks for
  multi-node representation learning.
\newblock In {\em Advances in Neural Information Processing Systems},
  volume~34, pages 9061--9073. Curran Associates, Inc., 2021.

\bibitem{DBLP:conf/nips/0007TYL19}
Shuai Zhang, Yi~Tay, Lina Yao, and Qi~Liu.
\newblock Quaternion knowledge graph embeddings.
\newblock In {\em Advances in Neural Information Processing Systems 32: Annual
  Conference on Neural Information Processing Systems 2019, NeurIPS 2019},
  pages 2731--2741, 2019.

\bibitem{zhang2021cone}
Zhanqiu Zhang, Jie Wang, Jiajun Chen, Shuiwang Ji, and Feng Wu.
\newblock Cone: Cone embeddings for multi-hop reasoning over knowledge graphs.
\newblock {\em Advances in Neural Information Processing Systems}, 34, 2021.

\bibitem{zhou2022ood}
Yangze Zhou, Gitta Kutyniok, and Bruno Ribeiro.
\newblock Ood link prediction generalization capabilities of message-passing
  gnns in larger test graphs.
\newblock In {\em Advances in Neural Information Processing Systems}, 2022.

\bibitem{gnn_qe}
Zhaocheng Zhu, Mikhail Galkin, Zuobai Zhang, and Jian Tang.
\newblock Neural-symbolic models for logical queries on knowledge graphs.
\newblock In {\em International Conference on Machine Learning, {ICML} 2022},
  Proceedings of Machine Learning Research. {PMLR}, 2022.

\bibitem{zhu2022torchdrug}
Zhaocheng Zhu, Chence Shi, Zuobai Zhang, Shengchao Liu, Minghao Xu, Xinyu Yuan,
  Yangtian Zhang, Junkun Chen, Huiyu Cai, Jiarui Lu, Chang Ma, Runcheng Liu,
  Louis-Pascal Xhonneux, Meng Qu, and Jian Tang.
\newblock Torchdrug: A powerful and flexible machine learning platform for drug
  discovery, 2022.

\bibitem{zhu2021neural}
Zhaocheng Zhu, Zuobai Zhang, Louis-Pascal Xhonneux, and Jian Tang.
\newblock Neural bellman-ford networks: A general graph neural network
  framework for link prediction.
\newblock {\em Advances in Neural Information Processing Systems}, 34, 2021.

\end{thebibliography}
\bibliographystyle{plain}

%\iffalse
\clearpage
%%%%%%%%%%%%%%%%%%%%%%%%%%%%%%%%%%%%%%%%%%%%%%%%%%%%%%%%%%%%
\section*{Checklist}

%%% BEGIN INSTRUCTIONS %%%
% The checklist follows the references.  Please
% read the checklist guidelines carefully for information on how to answer these
% questions.  For each question, change the default \answerTODO{} to \answerYes{},
% \answerNo{}, or \answerNA{}.  You are strongly encouraged to include a {\bf
% justification to your answer}, either by referencing the appropriate section of
% your paper or providing a brief inline description.  For example:
% \begin{itemize}
%   \item Did you include the license to the code and datasets? \answerYes{See Section~\ref{gen_inst}.}
%   \item Did you include the license to the code and datasets? \answerNo{The code and the data are proprietary.}
%   \item Did you include the license to the code and datasets? \answerNA{}
% \end{itemize}
% Please do not modify the questions and only use the provided macros for your
% answers.  Note that the Checklist section does not count towards the page
% limit.  In your paper, please delete this instructions block and only keep the
% Checklist section heading above along with the questions/answers below.
%%% END INSTRUCTIONS %%%

\begin{enumerate}

\item For all authors...
\begin{enumerate}
  \item Do the main claims made in the abstract and introduction accurately reflect the paper's contributions and scope?
    \answerYes{}
  \item Did you describe the limitations of your work?
    \answerYes{See Section~\ref{sec:conclusion}}
  \item Did you discuss any potential negative societal impacts of your work?
    \answerYes{See Section~\ref{sec:conclusion}}
  \item Have you read the ethics review guidelines and ensured that your paper conforms to them?
    \answerYes{}
\end{enumerate}

\item If you are including theoretical results...
\begin{enumerate}
  \item Did you state the full set of assumptions of all theoretical results?
    \answerNA{}
        \item Did you include complete proofs of all theoretical results?
    \answerNA{}
\end{enumerate}

\item If you ran experiments...
\begin{enumerate}
  \item Did you include the code, data, and instructions needed to reproduce the main experimental results (either in the supplemental material or as a URL)?
    \answerYes{Code and sample data are included in the supplementary material}
  \item Did you specify all the training details (e.g., data splits, hyperparameters, how they were chosen)?
    \answerYes{Dataset creation process is described in Section~\ref{sec:dataset} with more details in Appendix~\ref{app:dataset}. Hyperparameters are specified in Appendix~\ref{app:hyperparams}.}
        \item Did you report error bars (e.g., with respect to the random seed after running experiments multiple times)?
    \answerNo{We observe negligible variance w.r.t. random seeds}
        \item Did you include the total amount of compute and the type of resources used (e.g., type of GPUs, internal cluster, or cloud provider)?
    \answerYes{Training details are specified in Section~\ref{sec:dataset} and in Appendix~\ref{app:hyperparams}.}
\end{enumerate}

\item If you are using existing assets (e.g., code, data, models) or curating/releasing new assets...
\begin{enumerate}
  \item If your work uses existing assets, did you cite the creators?
    \answerYes{}
  \item Did you mention the license of the assets?
    \answerYes{}
  \item Did you include any new assets either in the supplemental material or as a URL?
    \answerYes{Due to the overall size, we include a sample of the benchmarking suite in the supplemental material and will openly publish the whole dataset.}
  \item Did you discuss whether and how consent was obtained from people whose data you're using/curating?
    \answerNA{No personal data involved}
  \item Did you discuss whether the data you are using/curating contains personally identifiable information or offensive content?
    \answerYes{The datasets are anonymized, we discuss it in Appendix~\ref{app:dataset}.}
\end{enumerate}

\item If you used crowdsourcing or conducted research with human subjects...
\begin{enumerate}
  \item Did you include the full text of instructions given to participants and screenshots, if applicable?
    \answerNA{}
  \item Did you describe any potential participant risks, with links to Institutional Review Board (IRB) approvals, if applicable?
    \answerNA{}
  \item Did you include the estimated hourly wage paid to participants and the total amount spent on participant compensation?
    \answerNA{}
\end{enumerate}

\end{enumerate}

%%%%%%%%%%%%%%%%%%%%%%%%%%%%%%%%%%%%%%%%%%%%%%%%%%%%%%%%%%%%
%\fi

\newpage
\appendix

\section{Differentiable Logical Operators}
\label{app:tnorms}

T-norms ($\top$) and t-conorms ($\bot$) are \emph{fuzzy} versions of conjunction ($\land$) and disjunction ($\lor$), respectively. 
Fuzzy operators can be applied to vectors of continuous values within a certain range, e.g., $[0, 1]^d$, depending on the chosen \emph{fuzzy logic}, and are executed as algebraic operations which makes them differentiable.
Different \emph{fuzzy logics} implement different t-norms and t-conorms. 
In this work, we experiment with two such logics: \emph{product logic} and \emph{G\"odel (min) logic}.
In the product logic, conjunction $\gC$, disjunction $\gD$, and negation $\gN$ are modeled as follows:

\begin{align*}
    \gC(\vx, \vy) &= \vx \odot \vy \\
    \gD(\vx, \vy) &= \vx + \vy - \vx \odot \vy \\
    \gN(\vx) &= \bm{1} - \vx
\end{align*}

where inputs $\vx, \vy \in [0,1]^d$ are $d$-dimensional vectors with values in the range $[0,1]$, $\odot$ is the element-wise multiplication, and $\bm{1}$ is the \emph{universe} vector of all ones.

In the G\"odel logic, conjunction $\gC$ and disjunction $\gD$ are modeled as \emph{min} and \emph{max}, respectively:

\begin{align*}
    \gC(\vx, \vy) &= \textit{min}(\vx, \vy) \\
    \gD(\vx, \vy) &= \textit{max}(\vx , \vy ) \\
\end{align*}

For GNN-QE we employ solely the product logic for end-to-end training on all types of complex queries. 
For NodePiece-QE and its inference-only mechanism based on CQD-Beam, we may select the best performing logic for each query type based on the validation set. 
The chosen operators for NodePiece-QE are reported in Table~\ref{tab:cqd_hparams} in Appedix~\ref{app:hyperparams}.

\section{Benchmarking Datasets Details}
\label{app:dataset}

We sampled 9 datasets (used in Section~\ref{sec:test_queries} and Section~\ref{sec:train_queries}) from the original FB15k-237~\cite{toutanova-chen-2015-observed} with already added inverse edges for ensuring reachability and connectedness of the underlying graph for the subsequent query sampling.
Creation details are provided in the Section~\ref{sec:dataset} and statistics on the sampled graphs are presented in Table~\ref{tab:graphs}. 
Varying the ratio of entities in the inference graph to the training graph $\gE_{\textit{inf}} / \gE_{\textit{train}}$, we aim at measuring inductive capabilities of proposed strategies in the out-of-distribution size generalization scenario. 
To measure scalability of inductive query answering approaches, we create WikiKG-QE, an inductive split of the originally transductive OGB WikiKG 2~\cite{DBLP:conf/nips/HuFZDRLCL20}, following the same sampling strategy as for 9 Freebase datasets. 

\begin{table}[!ht]
\centering
\caption{Sampled graphs statistics for various ratios $\gE_{\textit{inf}} / \gE_{\textit{train}}$. Originally inverse triples are included in all graphs except WikiKG-QE. $\gR$ - number of unique relation types, $\gE$ - number of entities in various splits, $\gT$ - number of triples. Validation and Test splits contain an inference graph $(\gE_\textit{inf}, \gT_\textit{inf})$ which is a superset of the training graph with new nodes, and missing edges to predict $\gT_\textit{pred}$.  }
\label{tab:graphs}
%\scriptsize
\begin{adjustbox}{width=\textwidth}
\begin{tabular}{cccccccccccc}\toprule
\multirow{2}{*}{Ratio, \%} &\multirow{2}{*}{$\gR$} &\multirow{2}{*}{$\gE_{\textit{total}}$} &\multicolumn{2}{c}{Training Graph} &\multicolumn{3}{c}{Validation Graph} &\multicolumn{3}{c}{Test Graph} \\ \cmidrule(lr){4-5}  \cmidrule(lr){6-8}  \cmidrule(lr){9-11}
& & &$\gE_{\textit{train}}$ & $\gT_{\textit{train}}$ & $\gE^{\textit{val}}_{\textit{inf}}$ & $\gT^{\textit{val}}_{\textit{inf}}$ & $\gT^{\textit{val}}_{\textit{pred}}$ & $\gE^{\textit{test}}_{\textit{inf}}$ & $\gT^{\textit{test}}_{\textit{inf}}$ & $\gT^{\textit{test}}_{\textit{pred}}$ \\ \midrule
%Ratio, \% & $\gR$ & $\gE_{\textit{total}}$ & $\gE_{\textit{train}}$ & $\gT_{\textit{train}}$ & $\gE^{\textit{val}}_{\textit{inf}}$ & $\gT^{\textit{val}}_{\textit{inf}}$ & $\gT^{\textit{val}}_{\textit{pred}}$ & $\gE^{\textit{test}}_{\textit{inf}}$ & $\gT^{\textit{test}}_{\textit{inf}}$ & $\gT^{\textit{test}}_{\textit{pred}}$ \\\midrule
106\% &466 &14,512 &13,091 &493,425 &13,801 &551,336 &10,219 &13,802 &538,896 &8,023 \\
113\% &468 &14,442 &11,601 &401,677 &13,022 &491,518 &15,849 &13,021 &486,068 &14,893 \\
122\% &466 &14,444 &10,184 &298,879 &12,314 &413,554 &20,231 &12,314 &430,892 &23,289 \\
134\% &466 &14,305 &8,634 &228,729 &11,468 &373,262 &25,477 &11,471 &367,810 &24,529 \\
150\% &462 &14,333 &7,232 &162,683 &10,783 &311,462 &26,235 &10,782 &331,352 &29,755 \\
175\% &436 &14,022 &5,560 &102,521 &9,801 &265,412 &28,691 &9,781 &266,494 &28,891 \\
217\% &446 &13,986 &4,134 &52,455 &9,062 &227,284 &30,809 &9,058 &212,386 &28,177 \\
300\% &412 &13,868 &2,650 &24,439 &8,252 &178,680 &27,135 &8,266 &187,156 &28,657 \\
550\% &312 &13,438 &1,084 &5,265 &7,247 &136,558 &22,981 &7,275 &133,524 &22,503 \\
\midrule
\midrule
        \multicolumn{11}{c}{\emph{WikiKG-QE}} \\ \midrule
133\% & 512 & 2,492,122 & 1,494,033 & 5,824,868 & 1,992,739  & 9,466,319 & 638,389 & 1,993,416 & 10,510,906 & 824,713 \\
\bottomrule
\end{tabular}
\end{adjustbox}
\end{table}

In all datasets, entities and relations are anonymized and only have an integer ID. Furthermode, inference graphs at validation and test time are supersets of the respective training graph with new nodes and edges.
The amount of new unique nodes is simply the difference $\gE_\textit{inf} - \gE_\textit{train}$ between entities in those graphs, e.g., for the dataset of ratio $175\%$, the validation inference graph contains $4,241$ new nodes and test inference graph contains $4,221$ news nodes. Note that those $4,241$ and $4,221$ nodes are unique for each graph and do not overlap. That is, validation inference and test inference graphs are disconnected except sharing the same core training graph.

Then, for each created inductive dataset, we sample queries of 14 query patterns following the BetaE~\cite{ren2020beta} procedure.
That is, \emph{training} queries are sampled from the \emph{training} graph $\gG_\textit{train}$ and have only \emph{easy} answers reachable by simple edge traversal. 
Validation and test queries are sampled from the respective splits, e.g., \emph{validation} queries are sampled from the validation graph $\gG_\textit{val}$ using entities from the validation inference graph $\gE^{\textit{val}}_\textit{inf}$ (which, in turn, are a union of training nodes and new, unseen validation nodes $\gE_\textit{train} \cup \gE_\textit{val}$), and at least one edge in each query belongs to $\gT^{\textit{val}}_{\textit{pred}}$ and has to be predicted during query execution.
Queries might have \emph{easy} answers that are directly reachable by traversing edges $\gT^{\textit{val}}_{\textit{inf}}$ in the validation inference graph, whereas \emph{hard} answers are only reachable after predicting missing edges from the set $\gT^{\textit{val}}_{\textit{pred}}$. 
Final evaluation metrics are computed only based on the \emph{hard} answers.
Following the literature~\cite{ren2020beta}, we only retain queries that have less than 1000 answers.
Table~\ref{tab:all_queries} summarizes the statistics on the sampled queries for each dataset ratio, each graph, and query type that we use in Section~\ref{sec:test_queries} for evaluating inductive query answering performance. 
In graphs with smaller inference graphs and smaller number of missing triples, we sample fewer queries with negation (\emph{2in, 3in, inp, pin, pni}) for validation and test splits.
For WikiKG-QE, due to its size, we only sample 10k EPFO queries to be executed in the inference-only regime without training (at the moment, CQD-Beam does not support queries with negation). 
We use those queries in Section~\ref{sec:scaling} to evaluate scalability of NodePiece-QE and prediction quality in the inference-only mode.

\begin{table}[!htp]
\centering
\caption{Statistics on sampled queries for each dataset ratio and query type. For WikiKG-QE, we only sample EPFO queries without negation.}
\label{tab:all_queries}
%\scriptsize
\begin{adjustbox}{width=\textwidth}
\begin{tabular}{lrrrrrrrrrrrrrrrr}\toprule
Ratio & Graph & \multicolumn{1}{c}{\textbf{1p}} & \multicolumn{1}{c}{\textbf{2p}} & \multicolumn{1}{c}{\textbf{3p}} & \multicolumn{1}{c}{\textbf{2i}} & \multicolumn{1}{c}{\textbf{3i}} & \multicolumn{1}{c}{\textbf{pi}} & \multicolumn{1}{c}{\textbf{ip}} & \multicolumn{1}{c}{\textbf{2u}} & \multicolumn{1}{c}{\textbf{up}} & \multicolumn{1}{c}{\textbf{2in}} & \multicolumn{1}{c}{\textbf{3in}} & \multicolumn{1}{c}{\textbf{inp}} & \multicolumn{1}{c}{\textbf{pin}} & \multicolumn{1}{c}{\textbf{pni}} \\\midrule
\multirow{3}{*}{106\%} &training & 135,613 &50,000 &50,000 &50,000 &50,000 &50,000 &50,000 &50,000 &50,000 &50,000 &40,000 &50,000 &50,000 &50,000 \\
&validation & 6,582 &10,000 &10,000 &10,000 &10,000 &10,000 &10,000 &10,000 &10,000 &1,000 &1,000 &1,000 &1,000 &1,000 \\
&test & 5,446 &10,000 &10,000 &10,000 &10,000 &10,000 &10,000 &10,000 &10,000 &1,000 &1,000 &1,000 &1,000 &1,000 \\ \midrule
\multirow{3}{*}{113\%} &training & 115,523 &50,000 &50,000 &50,000 &50,000 &50,000 &50,000 &50,000 &50,000 &50,000 &40,000 &50,000 &50,000 &50,000 \\
&validation & 10,256 &10,000 &10,000 &10,000 &10,000 &10,000 &10,000 &10,000 &10,000 &1,000 &1,000 &1,000 &1,000 &1,000 \\
&test & 9,782 &10,000 &10,000 &10,000 &10,000 &10,000 &10,000 &10,000 &10,000 &1,000 &1,000 &1,000 &1,000 &1,000 \\ \midrule
\multirow{3}{*}{122\%} &training & 91,228 &50,000 &50,000 &50,000 &50,000 &50,000 &50,000 &50,000 &50,000 &50,000 &40,000 &50,000 &50,000 &50,000 \\
&validation & 12,696 &10,000 &10,000 &10,000 &10,000 &10,000 &10,000 &10,000 &10,000 &5,000 &5,000 &5,000 &5,000 &5,000 \\
&test & 14,458 &10,000 &10,000 &10,000 &10,000 &10,000 &10,000 &10,000 &10,000 &5,000 &5,000 &5,000 &5,000 &5,000 \\ \midrule
\multirow{3}{*}{134\%} &training & 75,326 &50,000 &50,000 &50,000 &50,000 &50,000 &50,000 &50,000 &50,000 &50,000 &40,000 &50,000 &50,000 &50,000 \\
&validation & 15,541 &50,000 &50,000 &50,000 &50,000 &50,000 &50,000 &20,000 &20,000 &5,000 &5,000 &5,000 &5,000 &5,000 \\
&test & 15,270 &50,000 &50,000 &50,000 &50,000 &50,000 &50,000 &20,000 &20,000 &5,000 &5,000 &5,000 &5,000 &5,000 \\ \midrule
\multirow{3}{*}{150\%} &training & 56,114 &50,000 &50,000 &50,000 &50,000 &50,000 &50,000 &50,000 &50,000 &50,000 &40,000 &50,000 &50,000 &50,000 \\
&validation & 16,229 &50,000 &50,000 &50,000 &50,000 &50,000 &50,000 &50,000 &50,000 &5,000 &5,000 &5,000 &5,000 &5,000 \\
&test & 17,683 &50,000 &50,000 &50,000 &50,000 &50,000 &50,000 &50,000 &50,000 &5,000 &5,000 &5,000 &5,000 &5,000 \\ \midrule
\multirow{3}{*}{175\%} &training & 38,851 &50,000 &50,000 &50,000 &50,000 &50,000 &50,000 &50,000 &50,000 &50,000 &40,000 &50,000 &50,000 &50,000 \\
&validation & 17,235 &50,000 &50,000 &50,000 &50,000 &50,000 &50,000 &50,000 &50,000 &10,000 &10,000 &10,000 &10,000 &10,000 \\
&test & 17,476 &50,000 &50,000 &50,000 &50,000 &50,000 &50,000 &50,000 &50,000 &10,000 &10,000 &10,000 &10,000 &10,000 \\ \midrule
\multirow{3}{*}{217\%} & training & 22,422 &30,000 &30,000 &50,000 &50,000 &50,000 &50,000 &50,000 &50,000 &30,000 &30,000 &50,000 &50,000 &50,000 \\
&validation & 18,168 &50,000 &50,000 &50,000 &50,000 &50,000 &50,000 &50,000 &50,000 &10,000 &10,000 &10,000 &10,000 &10,000 \\
&test & 16,902 &50,000 &50,000 &50,000 &50,000 &50,000 &50,000 &50,000 &50,000 &10,000 &10,000 &10,000 &10,000 &10,000 \\ \midrule
\multirow{3}{*}{300\%} &training & 11,699 &15,000 &15,000 &40,000 &40,000 &50,000 &50,000 &50,000 &50,000 &15,000 &15,000 &50,000 &40,000 &50,000 \\
&validation & 16,189 &50,000 &50,000 &50,000 &50,000 &50,000 &50,000 &50,000 &50,000 &10,000 &10,000 &10,000 &10,000 &10,000 \\
&test & 17,105 &50,000 &50,000 &50,000 &50,000 &50,000 &50,000 &50,000 &50,000 &10,000 &10,000 &10,000 &10,000 &10,000 \\ \midrule
\multirow{3}{*}{550\%} &training & 3,284 &15,000 &15,000 &40,000 &40,000 &50,000 &50,000 &50,000 &50,000 &10,000 &10,000 &30,000 &30,000 &30,000 \\
&validation & 13,616 &50,000 &50,000 &50,000 &50,000 &50,000 &50,000 &50,000 &50,000 &10,000 &10,000 &10,000 &10,000 &10,000 \\
&test & 13,670 &50,000 &50,000 &50,000 &50,000 &50,000 &50,000 &50,000 &50,000 &10,000 &10,000 &10,000 &10,000 &10,000 \\ 
\midrule
\midrule
        \multicolumn{16}{c}{\emph{WikiKG-QE}} \\ \midrule
\multirow{3}{*}{133\%} & training & 10,000 & 10,000 & 10,000 & 10,000 & 10,000 & 10,000 & 10,000 & 10,000 & 10,000 & - & - & - & - & -  \\        &validation & 10,000 & 10,000 & 10,000 & 10,000 & 10,000 & 10,000 & 10,000 & 10,000 & 10,000 & - & - & - & - & - \\
&test & 10,000 & 10,000 & 10,000 & 10,000 & 10,000 & 10,000 & 10,000 & 10,000 & 10,000 & - & - & - & - & - \\ 
\bottomrule
\end{tabular}
\end{adjustbox}
\end{table}

\begin{table}[!htp]
\centering
\caption{Statistics on \textbf{training} EPFO queries that have a different (often, larger) answer set when executed against validation and test inference graphs. We list the original number of training queries, number of those queries with new \emph{easy} answers in the validation (In val) and test graphs (In test), as well as their percentage ratio to the total number. Most queries (except \emph{2i,3i}) have new answer sets.}
\label{tab:train_queries_epfo}
%\scriptsize
\begin{adjustbox}{width=\textwidth}
\begin{tabular}{lrrrrrrrrrrrrrrrrrrrr}\toprule
\multirow{2}{*}{\textbf{Ratio}} &\multirow{2}{*}{\textbf{Graph}} &\multicolumn{2}{c}{\textbf{1p}} &\multicolumn{2}{c}{\textbf{2p}} &\multicolumn{2}{c}{\textbf{3p}} &\multicolumn{2}{c}{\textbf{2i}} &\multicolumn{2}{c}{\textbf{3i}} &\multicolumn{2}{c}{\textbf{pi}} &\multicolumn{2}{c}{\textbf{ip}} &\multicolumn{2}{c}{\textbf{2u}} &\multicolumn{2}{c}{\textbf{up}} \\ \cmidrule(lr){3-4} \cmidrule(lr){5-6} \cmidrule(lr){7-8} \cmidrule(lr){9-10} \cmidrule(lr){11-12} \cmidrule(lr){13-14} \cmidrule(lr){15-16} \cmidrule(lr){17-18} \cmidrule(lr){19-20}
& &\#Q &\% &\#Q &\% &\#Q &\% &\#Q &\% &\#Q &\% &\#Q &\% &\#Q &\% &\#Q &\% &\#Q &\% \\ \midrule
%\textbf{Ratio} &\textbf{Queries} &\multicolumn{2}{c}{\textbf{1p (\%)}} &\multicolumn{2}{c}{\textbf{2p}} &\multicolumn{2}{c}{\textbf{3p}} &\multicolumn{2}{c}{\textbf{2i}} &\multicolumn{2}{c}{\textbf{3i}} &\multicolumn{2}{c}{\textbf{pi}} &\multicolumn{2}{c}{\textbf{ip}} &\multicolumn{2}{c}{\textbf{2u}} &\multicolumn{2}{c}{\textbf{up}} \\\midrule
\multirow{3}{*}{106\%} &Train &135,613 &100.0 &50,000 &100.0 &50,000 &100.0 &50,000 &100.0 &50,000 &100.0 &50,000 &100.0 &50,000 &100.0 &50,000 &100.0 &50,000 &100.0 \\
&In val &14,079 &10.4 &32,220 &64.4 &40,860 &81.7 &7,598 &15.2 &4,416 &8.8 &16,485 &33.0 &29,290 &58.6 &33,507 &67.0 &41,671 &83.3 \\
&In test &11,560 &8.5 &31,894 &63.8 &40,547 &81.1 &7,313 &14.6 &4,175 &8.4 &16,204 &32.4 &28,778 &57.6 &32,978 &66.0 &41,167 &82.3 \\ \midrule
\multirow{3}{*}{113\%} &Train &115,523 &100.0 &50,000 &100.0 &50,000 &100.0 &50,000 &100.0 &50,000 &100.0 &50,000 &100.0 &50,000 &100.0 &50,000 &100.0 &50,000 &100.0 \\
&In val &17,792 &15.4 &36,499 &73.0 &43,473 &86.9 &10,517 &21.0 &6,394 &12.8 &20,556 &41.1 &33,599 &67.2 &37,955 &75.9 &44,318 &88.6 \\
&In test &17,576 &15.2 &36,721 &73.4 &43,541 &87.1 &10,552 &21.1 &6,303 &12.6 &20,382 &40.8 &33,726 &67.5 &38,107 &76.2 &44,501 &89.0 \\ \midrule
\multirow{3}{*}{122\%} &Train &91,228 &100.0 &50,000 &100.0 &50,000 &100.0 &50,000 &100.0 &50,000 &100.0 &50,000 &100.0 &50,000 &100.0 &50,000 &100.0 &50,000 &100.0 \\
&In val &20,281 &22.2 &38,642 &77.3 &44,654 &89.3 &11,695 &23.4 &5,851 &11.7 &22,662 &45.3 &35,935 &71.9 &40,356 &80.7 &45,672 &91.3 \\
&In test &20,418 &22.4 &38,706 &77.4 &44,688 &89.4 &11,847 &23.7 &6,185 &12.4 &22,524 &45.0 &35,768 &71.5 &40,395 &80.8 &45,684 &91.4 \\ \midrule
\multirow{3}{*}{134\%} &Train &75,326 &100.0 &50,000 &100.0 &50,000 &100.0 &50,000 &100.0 &50,000 &100.0 &50,000 &100.0 &50,000 &100.0 &50,000 &100.0 &50,000 &100.0 \\
&In val &18,909 &25.1 &39,893 &79.8 &45,253 &90.5 &14,256 &28.5 &8,655 &17.3 &24,619 &49.2 &37,835 &75.7 &41,899 &83.8 &46,114 &92.2 \\
&In test &19,372 &25.7 &39,762 &79.5 &45,325 &90.7 &14,082 &28.2 &8,790 &17.6 &24,212 &48.4 &37,527 &75.1 &41,494 &83.0 &46,210 &92.4 \\ \midrule
\multirow{3}{*}{150\%} &Train &56,114 &100.0 &50,000 &100.0 &50,000 &100.0 &50,000 &100.0 &50,000 &100.0 &50,000 &100.0 &50,000 &100.0 &50,000 &100.0 &50,000 &100.0 \\
&In val &17,434 &31.1 &40,666 &81.3 &45,832 &91.7 &14,103 &28.2 &8,011 &16.0 &25,106 &50.2 &38,499 &77.0 &42,587 &85.2 &46,754 &93.5 \\
&In test &18,566 &33.1 &41,202 &82.4 &46,092 &92.2 &14,575 &29.2 &8,193 &16.4 &25,782 &51.6 &38,932 &77.9 &43,101 &86.2 &46,791 &93.6 \\ \midrule
\multirow{3}{*}{175\%} &Train &38,851 &100.0 &50,000 &100.0 &50,000 &100.0 &50,000 &100.0 &50,000 &100.0 &50,000 &100.0 &50,000 &100.0 &50,000 &100.0 &50,000 &100.0 \\
&In val &14,063 &36.2 &41,290 &82.6 &46,214 &92.4 &15,645 &31.3 &9,222 &18.4 &27,205 &54.4 &40,161 &80.3 &44,128 &88.3 &47,366 &94.7 \\
&In test &14,214 &36.6 &41,143 &82.3 &46,061 &92.1 &15,731 &31.5 &9,391 &18.8 &27,207 &54.4 &40,297 &80.6 &43,980 &88.0 &47,319 &94.6 \\ \midrule
\multirow{3}{*}{217\%} &Train &22,422 &100.0 &30,000 &100.0 &30,000 &100.0 &50,000 &100.0 &50,000 &100.0 &50,000 &100.0 &50,000 &100.0 &50,000 &100.0 &50,000 &100.0 \\
&In val &10,437 &46.5 &24,659 &82.2 &26,760 &89.2 &13,784 &27.6 &7,807 &15.6 &24,884 &49.8 &39,107 &78.2 &43,496 &87.0 &46,112 &92.2 \\
&In test &10,257 &45.7 &24,344 &81.1 &26,579 &88.6 &14,055 &28.1 &7,962 &15.9 &24,962 &49.9 &38,966 &77.9 &43,092 &86.2 &45,850 &91.7 \\ \midrule
\multirow{3}{*}{300\%} &Train &11,699 &100.0 &15,000 &100.0 &15,000 &100.0 &40,000 &100.0 &40,000 &100.0 &50,000 &100.0 &50,000 &100.0 &50,000 &100.0 &50,000 &100.0 \\
&In val &5,830 &49.8 &12,366 &82.4 &13,230 &88.2 &12,833 &32.1 &7,911 &19.8 &27,920 &55.8 &40,800 &81.6 &43,516 &87.0 &46,453 &92.9 \\
&In test &6,061 &51.8 &12,477 &83.2 &13,309 &88.7 &13,291 &33.2 &8,284 &20.7 &28,447 &56.9 &41,214 &82.4 &43,966 &87.9 &46,668 &93.3 \\ \midrule
\multirow{3}{*}{550\%} &Train &3,284 &100.0 &15,000 &100.0 &15,000 &100.0 &40,000 &100.0 &40,000 &100.0 &50,000 &100.0 &50,000 &100.0 &50,000 &100.0 &50,000 &100.0 \\
&In val &1,885 &57.4 &11,484 &76.6 &12,575 &83.8 &11,119 &27.8 &6,617 &16.5 &23,126 &46.3 &39,243 &78.5 &38,129 &76.3 &45,173 &90.3 \\
&In test &1,883 &57.3 &11,597 &77.3 &12,654 &84.4 &11,244 &28.1 &6,795 &17.0 &23,575 &47.2 &39,630 &79.3 &37,508 &75.0 &45,412 &90.8 \\
\bottomrule
\end{tabular}
\end{adjustbox}
\end{table}

\begin{table}[!htp]\centering
\caption{Statistics on \textbf{training} negation queries that have a different (often, larger) answer set when executed against validation and test inference graphs. We list the original number of training queries, number of those queries with new \emph{easy} answers in the validation (In val) and test graphs (In test), as well as their percentage ratio to the total number. Most queries have new answer sets.}
\label{tab:train_queries_neg}
%\scriptsize
\begin{adjustbox}{width=\textwidth}
\begin{tabular}{lrrrrrrrrrrrr}\toprule
\multirow{2}{*}{\textbf{Ratio}} &\multirow{2}{*}{\textbf{Graph}} &\multicolumn{2}{c}{\textbf{2in}} &\multicolumn{2}{c}{\textbf{3in}} &\multicolumn{2}{c}{\textbf{pin}} &\multicolumn{2}{c}{\textbf{pni}} &\multicolumn{2}{c}{\textbf{inp}} \\ \cmidrule(lr){3-4} \cmidrule(lr){5-6} \cmidrule(lr){7-8} \cmidrule(lr){9-10} \cmidrule(lr){11-12}
& &\#Q &\% &\#Q &\% &\#Q &\% &\#Q &\% &\#Q &\% \\\midrule
\multirow{3}{*}{106\%} &Train &50,000 &100.0 &50,000 &100.0 &50,000 &100.0 &50,000 &100.0 &50,000 &100.0 \\
&In val & 25,318 &50.6 &18,232 &36.5 &37,857 &75.7 &27,572 &55.1 &37,497 &75.0 \\
&In test & 25,111 &50.2 &18,237 &36.5 &37,441 &74.9 &27,535 &55.1 &37,176 &74.4 \\  \midrule
\multirow{3}{*}{113\%} &Train &50,000 &100.0 &50,000 &100.0 &50,000 &100.0 &50,000 &100.0 &50,000 &100.0 \\
&In val &31,216 &62.4 &24,620 &49.2 &42,015 &84.0 &33,011 &66.0 &41,980 &84.0 \\
&In test &31,437 &62.9 &24,665 &49.3 &42,255 &84.5 &33,115 &66.2 &42,296 &84.6 \\  \midrule
\multirow{3}{*}{122\%} &Train &50,000 &100.0 &50,000 &100.0 &50,000 &100.0 &50,000 &100.0 &50,000 &100.0 \\
&In val &34,722 &69.4 &26,700 &53.4 &44,104 &88.2 &36,361 &72.7 &44,070 &88.1 \\
&In test &35,028 &70.1 &27,105 &54.2 &44,089 &88.2 &36,398 &72.8 &44,074 &88.1 \\  \midrule
\multirow{3}{*}{134\%} &Train &50,000 &100.0 &50,000 &100.0 &50,000 &100.0 &50,000 &100.0 &50,000 &100.0 \\
&In val &38,096 &76.2 &31,631 &63.3 &45,672 &91.3 &39,641 &79.3 &45,491 &91.0 \\
&In test &37,469 &74.9 &31,224 &62.4 &45,521 &91.0 &38,971 &77.9 &45,418 &90.8 \\  \midrule
\multirow{3}{*}{150\%} &Train &50,000 &100.0 &40,000 &100.0 &50,000 &100.0 &50,000 &100.0 &50,000 &100.0 \\
&In val &39,836 &79.7 &26,534 &66.3 &46,561 &93.1 &40,733 &81.5 &46,496 &93.0 \\
&In test &40,127 &80.3 &26,968 &67.4 &46,832 &93.7 &41,100 &82.2 &46,811 &93.6 \\  \midrule
\multirow{3}{*}{175\%} &Train &50,000 &100.0 &40,000 &100.0 &50,000 &100.0 &50,000 &100.0 &50,000 &100.0 \\
&In val &42,418 &84.8 &29,083 &72.7 &47,666 &95.3 &42,987 &86.0 &47,606 &95.2 \\
&In test &42,379 &84.8 &29,170 &72.9 &47,749 &95.5 &42,941 &85.9 &47,557 &95.1 \\  \midrule
\multirow{3}{*}{217\%} &Train &30,000 &100.0 &30,000 &100.0 &50,000 &100.0 &50,000 &100.0 &50,000 &100.0 \\
&In val &26,202 &87.3 &21,751 &72.5 &47,879 &95.8 &43,958 &87.9 &47,688 &95.4 \\
&In test &26,080 &86.9 &21,591 &72.0 &47,655 &95.3 &43,837 &87.7 &47,417 &94.8 \\  \midrule
\multirow{3}{*}{300\%} &Train &15,000 &100.0 &15,000 &100.0 &50,000 &100.0 &40,000 &100.0 &50,000 &100.0 \\
&In val &13,595 &90.6 &11,996 &80.0 &48,693 &97.4 &36,427 &91.1 &48,279 &96.6 \\
&In test &13,659 &91.1 &12,098 &80.7 &48,791 &97.6 &36,507 &91.3 &48,440 &96.9 \\  \midrule
\multirow{3}{*}{550\%} &Train &10,000 &100.0 &10,000 &100.0 &30,000 &100.0 &30,000 &100.0 &30,000 &100.0 \\
&In val &9,232 &92.3 &8,071 &80.7 &29,484 &98.3 &27,975 &93.3 &29,393 &98.0 \\
&In test &9,137 &91.4 &8,053 &80.5 &29,510 &98.4 &27,839 &92.8 &29,218 &97.4 \\
\bottomrule
\end{tabular}
\end{adjustbox}
\end{table}

Furthermore, for the experiment in Section~\ref{sec:train_queries} to measure the abilities of inductive models to find new answers of known queries, we take the created \textbf{training} queries and find their \emph{easy} answers in the validation inference $\gG_\textit{inf}^\textit{val}=(\gE^{\textit{val}}_{\textit{inf}}, \gT^{\textit{val}}_{\textit{inf}})$ and test inference $\gG_\textit{inf}^\textit{test}=(\gE^{\textit{test}}_{\textit{inf}}, \gT^{\textit{test}}_{\textit{inf}})$ graphs. 
That is, those new answers do not require predicting missing edges in the inference graphs and only require a model to execute edge traversal to find (if any) new correct answers involving new, unseen entities and edges.
For the validation (test) split, we only count such training queries $q$ whose answer set in this split is \emph{different} from the answer set in the training graph, e.g., $\gA_q^{\textit{val}} \neq \gA_q^{\textit{train}}$.
We summarize the statistics of identified new answer sets in all datasets in Table~\ref{tab:train_queries_epfo} (for EPFO queries) and Table~\ref{tab:train_queries_neg} (for queries with negations).
We find that in most query patterns across all dataset ratios, training queries indeed have new answer sets when executed against validation or test inference graphs.

\section{More Experimental Results}
\label{app:more_experiments}

\begin{figure}[!ht]
    \centering
    \includegraphics[width=\textwidth]{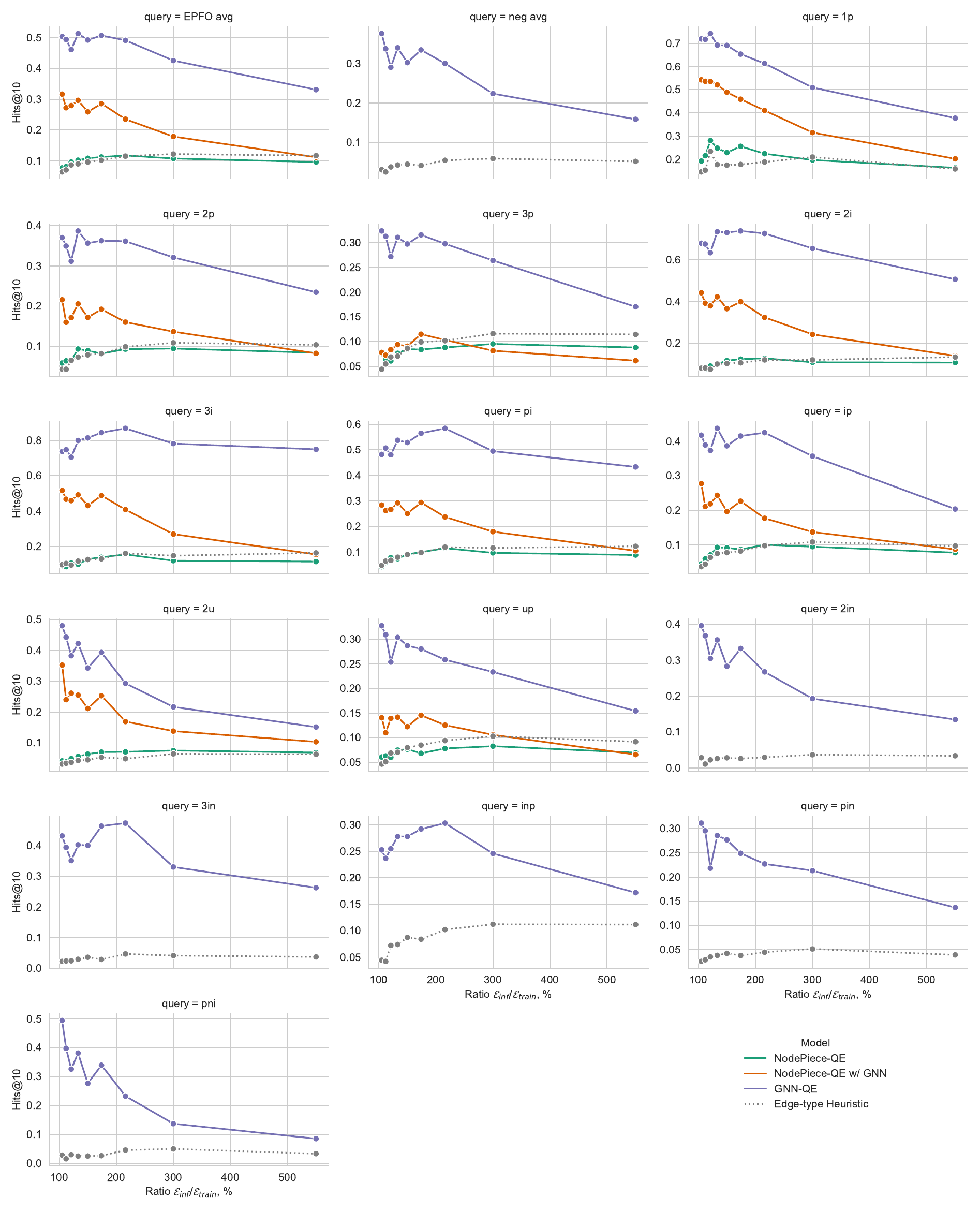}
    \caption{Hits@10 results on answering \textbf{test} inductive FOL queries on all ratios $\gE_{\textit{inf}} / \gE_{\textit{train}}$.}
    \label{fig:test_h10}
\end{figure}

Here, we present a detailed breakdown of query answering performance measured in Sections~\ref{sec:test_queries} and \ref{sec:train_queries} by query type. 
Fig.~\ref{fig:test_h10} and Table~\ref{tab:more_test_queries} contain detailed results from Section~\ref{sec:test_queries} of executing \textbf{test} queries with new, unseen entities over inference graphs of various ratios of new entities.

\begin{table}[t]
\caption{Test Hits@3 and Hits@10 results (\%) on answering \textbf{test} inductive FOL queries on all ratios $\gE_{\textit{inf}} / \gE_{\textit{train}}$. avg$_p$ is the average on EPFO queries ($\land$, $\lor$). avg$_n$ is the average on queries with negation.}
\label{tab:more_test_queries}
\begin{adjustbox}{width=\textwidth}
\begin{tabular}{@{}cllrrrrrrrrrrrrrrrr@{}}
\toprule
\multicolumn{1}{l}{Ratio} & Model & Metric & \bf{avg$_p$} & \bf{avg$_n$}  & 1p & 2p & 3p & 2i & 3i & pi & ip & 2u & up & 2in & 3in & inp & pin & pni \\ \midrule
\multirow{8}{*}{550\%} & \multirow{2}{*}{Edge-type Heuristic} & Hits@3 &5.0 &2.3 &5.9 &4.7 &5.2 &5.3 &7.0 &5.1 &4.3 &2.8 &4.5 &1.5 &1.7 &4.9 &2.0 &1.4 \\
 &  & Hits@10 &11.7 &5.1 &15.8 &10.4 &11.5 &13.4 &16.4 &12.2 &9.7 &6.3 &9.2 &3.4 &3.7 &11.2 &3.9 &3.4 \\ \cmidrule{2-19}
 & \multirow{2}{*}{NodePiece-QE} & Hits@3 &4.3 & &7.3 &4.0 &4.1 &4.3 &4.5 &3.8 &3.6 &3.4 &3.3 & & & & & \\
 &  & Hits@10 &9.6 & &16.3 &8.4 &8.8 &10.8 &11.5 &8.8 &7.7 &6.8 &7.0 & & & & &  \\ \cmidrule{2-19}
 & \multirow{2}{*}{NodePiece-QE w/ GNN} & Hits@3 &5.4 & &9.2 &4.1 &3.1 &6.8 &7.4 &5.1 &4.5 &5.4 &3.4 & & & & &  \\
 &  & Hits@10 &11.1 & &20.1 &8.3 &6.2 &14.0 &15.5 &10.5 &8.7 &10.3 &6.6 & & & & &  \\ \cmidrule{2-19}
 & \multirow{2}{*}{GNN-QE} & Hits@3 &24.2 &9.7 &28.3 &15.8 &10.1 &37.7 &60.9 &31.3 &14.4 &10.1 &8.8 &9.0 &16.5 &9.8 &7.7 &5.6 \\
 &  & Hits@10 &33.1 &15.8 &37.7 &23.4 &17.0 &50.7 &74.9 &43.3 &20.4 &15.1 &15.4 &13.4 &26.3 &17.2 &13.7 &8.6 \\ \midrule
\multirow{8}{*}{300\%} & \multirow{2}{*}{Edge-type Heuristic} & Hits@3 &5.5 &2.7 &10.3 &5.1 &5.4 &5.0 &6.3 &5.0 &4.8 &2.7 &5.1 &1.6 &1.9 &4.8 &2.5 &2.5 \\
 &  & Hits@10 &12.2 &5.8 &20.9 &10.9 &11.6 &12.1 &14.8 &11.6 &10.8 &6.4 &10.3 &3.6 &4.1 &11.2 &5.1 &5.0  \\ \cmidrule{2-19}
 & \multirow{2}{*}{NodePiece-QE} & Hits@3 &5.4 & &12.0 &4.7 &4.6 &4.9 &5.2 &4.2 &4.6 &3.9 &4.0 & & & & &  \\
 &  & Hits@10 &10.7 & &19.6 &9.5 &9.5 &11.0 &12.0 &9.7 &9.5 &7.5 &8.3 & & & & &  \\ \cmidrule{2-19}
 & \multirow{2}{*}{NodePiece-QE w/ GNN} & Hits@3 &9.7 & &18.9 &7.3 &4.1 &13.1 &15.0 &9.1 &7.3 &7.2 &5.5 & & & & & \\
 &  & Hits@10 &17.9 & &31.5 &13.7 &8.2 &24.3 &27.0 &18.0 &13.7 &13.8 &10.6 & & & & & \\ \cmidrule{2-19}
 & \multirow{2}{*}{GNN-QE} & Hits@3 &31.8 &13.5 &41.5 &21.0 &16.1 &51.7 &66.7 &37.2 &25.2 &13.5 &13.3 &11.7 &21.3 &14.0 &12.7 &7.7 \\
 &  & Hits@10 &42.6 &22.4 &50.9 &32.1 &26.4 &65.4 &78.1 &49.5 &35.7 &21.7 &23.4 &19.3 &33.1 &24.6 &21.3 &13.7 \\ \midrule
\multirow{8}{*}{217\%} & \multirow{2}{*}{Edge-type Heuristic} & Hits@3 &5.5 &2.4 &10.3 &4.8 &4.8 &5.1 &7.2 &5.4 &4.7 &2.1 &4.6 &1.2 &2.3 &4.3 &2.2 &2.2 \\
 &  & Hits@10 &11.5 &5.4 &18.8 &9.9 &10.2 &12.1 &16.1 &11.9 &9.8 &4.8 &9.4 &2.9 &4.7 &10.2 &4.5 &4.6 \\ \cmidrule{2-19}
 & \multirow{2}{*}{NodePiece-QE} & Hits@3 &5.9 & &13.9 &4.8 &4.4 &5.8 &6.7 &5.3 &5.1 &3.4 &4.0 & & & & & \\
 &  & Hits@10 &11.7 & &22.3 &9.3 &8.8 &12.9 &15.5 &11.5 &10.0 &7.1 &7.8 & & & & & \\ \cmidrule{2-19}
 & \multirow{2}{*}{NodePiece-QE w/ GNN} & Hits@3 &13.6 & &25.7 &8.8 &5.3 &18.7 &24.8 &12.9 &10.3 &8.9 &6.9 & & & & & \\
 &  & Hits@10 &23.5 & &41.0 &16.0 &10.4 &32.5 &40.8 &23.7 &17.7 &16.9 &12.6 & & & & & \\ \cmidrule{2-19}
 & \multirow{2}{*}{GNN-QE} & Hits@3 &37.9 &19.2 &50.6 &24.4 &19.3 &58.6 &76.2 &45.1 &31.4 &19.7 &16.0 &17.6 &32.6 &18.3 &14.0 &13.6 \\
 &  & Hits@10 &49.2 &30.1 &61.3 &36.1 &29.8 &72.6 &86.8 &58.4 &42.5 &29.3 &25.8 &26.7 &47.4 &30.3 &22.7 &23.3 \\ \midrule
\multirow{8}{*}{175\%} & \multirow{2}{*}{Edge-type Heuristic} & Hits@3 &4.7 &1.7 &8.4 &3.8 &4.8 &4.5 &5.6 &4.3 &4.0 &2.5 &4.2 &1.0 &1.2 &3.3 &1.8 &1.0 \\
 &  & Hits@10 &10.1 &4.1 &17.7 &8.2 &9.9 &10.7 &13.0 &9.8 &8.2 &5.3 &8.5 &2.6 &2.9 &8.4 &3.8 &2.7 \\ \cmidrule{2-19}
 & \multirow{2}{*}{NodePiece-QE} & Hits@3 &5.6 & &14.2 &4.1 &4.1 &5.6 &6.2 &4.5 &4.6 &3.5 &3.3 & & & & & \\
 &  & Hits@10 &11.2 & &25.5 &8.2 &8.4 &12.4 &13.9 &9.9 &8.7 &7.0 &6.8 & & & & & \\ \cmidrule{2-19}
 & \multirow{2}{*}{NodePiece-QE w/ GNN} & Hits@3 &17.2 & &30.7 &10.7 &5.9 &24.4 &31.2 &17.2 &13.1 &14.2 &7.3 & & & & & \\
 &  & Hits@10 &28.6 & &45.9 &19.2 &11.5 &39.9 &48.8 &29.4 &22.6 &25.3 &14.6 & & & & & \\ \cmidrule{2-19}
 & \multirow{2}{*}{GNN-QE} & Hits@3 &38.5 &20.5 &52.8 &24.1 &20.6 &59.8 &73.3 &43.2 &30.0 &24.4 &17.9 &18.9 &32.2 &17.8 &15.3 &18.2 \\
 &  & Hits@10 &50.7 &33.6 &65.4 &36.3 &31.6 &73.8 &84.3 &56.5 &41.5 &39.3 &28.0 &33.3 &46.4 &29.2 &24.9 &34.0 \\ \midrule
\multirow{8}{*}{150\%} & \multirow{2}{*}{Edge-type Heuristic} & Hits@3 &4.4 &1.9 &9.2 &3.6 &4.0 &4.3 &5.3 &3.9 &3.5 &1.8 &3.8 &1.3 &1.5 &3.5 &2.1 &1.1 \\
 &  & Hits@10 &9.6 &4.4 &17.4 &7.9 &8.7 &10.4 &12.7 &9.0 &7.7 &4.5 &8.0 &2.8 &3.6 &8.7 &4.2 &2.5 \\ \cmidrule{2-19}
 & \multirow{2}{*}{NodePiece-QE} & Hits@3 &5.4 & &14.0 &4.5 &4.1 &5.0 &5.5 &4.0 &4.6 &3.0 &3.8 & & & & & \\
 &  & Hits@10 &10.8 & &22.8 &8.9 &8.5 &11.7 &12.9 &9.1 &9.1 &6.3 &7.7 & & & & & \\ \cmidrule{2-19}
 & \multirow{2}{*}{NodePiece-QE w/ GNN} & Hits@3 &15.7 & &33.1 &9.7 &4.6 &22.3 &26.9 &14.8 &11.4 &12.3 &6.5 & & & & & \\
 &  & Hits@10 &25.9 & &48.9 &17.2 &9.1 &36.6 &43.1 &25.1 &19.7 &21.1 &12.2 & & & & & \\ \cmidrule{2-19}
 & \multirow{2}{*}{GNN-QE} & Hits@3 &37.3 &18.1 &56.6 &23.6 &18.9 &58.6 &69.8 &39.6 &27.3 &23.2 &18.0 &16.9 &25.7 &16.6 &16.2 &15.4 \\
 &  & Hits@10 &49.3 &30.3 &69.1 &35.7 &29.7 &73.1 &81.3 &52.9 &38.7 &34.3 &28.7 &28.3 &40.1 &27.8 &27.7 &27.7 \\ \midrule
\multirow{8}{*}{133\%} & \multirow{2}{*}{Edge-type Heuristic} & Hits@3 &4.0 &1.9 &8.6 &3.5 &3.2 &4.3 &4.9 &3.4 &3.5 &1.8 &3.2 &1.2 &1.6 &2.7 &2.0 &1.1 \\
 &  & Hits@10 &9.0 &4.2 &17.7 &7.3 &7.1 &10.1 &11.8 &8.0 &7.5 &4.3 &7.0 &2.6 &2.9 &7.4 &3.8 &2.5 \\ \cmidrule{2-19}
 & \multirow{2}{*}{NodePiece-QE} & Hits@3 &5.1 & &15.4 &4.8 &3.5 &4.4 &4.1 &2.9 &4.8 &2.6 &3.4 & & & & &  \\
 &  & Hits@10 &10.2 & &24.8 &9.3 &7.7 &10.1 &9.9 &7.4 &9.3 &5.6 &7.5 & & & & &  \\ \cmidrule{2-19}
 & \multirow{2}{*}{NodePiece-QE w/ GNN} & Hits@3 &19.4 & &38.0 &12.6 &5.2 &27.0 &32.3 &17.9 &16.0 &16.7 &8.7 & & & & & \\
 &  & Hits@10 &29.6 & &52.1 &20.6 &9.4 &42.3 &49.2 &29.3 &24.3 &25.5 &14.2 & & & & & \\ \cmidrule{2-19}
 & \multirow{2}{*}{GNN-QE} & Hits@3 &38.8 &21.4 &56.3 &25.6 &19.8 &59.3 &68.5 &40.6 &30.6 &28.4 &19.8 &23.0 &25.9 &16.4 &18.3 &23.6 \\
 &  & Hits@10 &51.4 &34.1 &69.2 &38.7 &31.1 &73.4 &79.9 &53.8 &43.7 &42.2 &30.4 &35.6 &40.3 &27.8 &28.6 &38.1 \\ \midrule
\multirow{8}{*}{121\%} & \multirow{2}{*}{Edge-type Heuristic} & Hits@3 &4.3 &1.5 &14.7 &3.0 &3.2 &3.0 &3.9 &2.8 &2.8 &1.5 &3.3 &0.9 &1.0 &2.6 &1.7 &1.2 \\
 &  & Hits@10 &8.6 &3.7 &23.3 &6.5 &6.9 &7.6 &9.5 &6.7 &6.4 &3.7 &6.9 &2.2 &2.4 &7.2 &3.5 &3.0 \\ \cmidrule{2-19}
 & \multirow{2}{*}{NodePiece-QE} & Hits@3 &4.6 & &16.0 &3.2 &2.7 &3.7 &4.3 &3.1 &3.5 &2.1 &2.8 & & & & & \\
 &  & Hits@10 &9.6 & &28.0 &6.5 &6.1 &9.2 &10.7 &7.8 &7.1 &4.9 &6.0 & & & & & \\ \cmidrule{2-19}
 & \multirow{2}{*}{NodePiece-QE w/ GNN} & Hits@3 &18.4 & &39.7 &10.6 &4.8 &24.8 &30.6 &16.4 &13.9 &16.8 &7.8 & & & & & \\
 &  & Hits@10 &27.9 & &53.6 &17.1 &8.4 &38.0 &45.9 &26.7 &21.9 &26.1 &13.9 & & & & & \\ \cmidrule{2-19}
 & \multirow{2}{*}{GNN-QE} & Hits@3 &35.3 &18.9 &62.0 &21.1 &17.9 &50.0 &59.5 &36.8 &26.7 &27.4 &16.6 &20.2 &23.4 &15.4 &13.8 &21.5 \\
 &  & Hits@10 &46.2 &29.1 &74.2 &31.1 &27.2 &63.4 &70.5 &48.1 &37.3 &38.3 &25.4 &30.5 &35.2 &25.5 &21.8 &32.6 \\ \midrule
\multirow{8}{*}{113\%} & \multirow{2}{*}{Edge-type Heuristic} & Hits@3 &3.1 &1.0 &8.5 &1.8 &2.3 &3.3 &4.5 &2.5 &1.9 &1.3 &2.1 &0.5 &1.3 &1.5 &1.2 &0.4 \\
 &  & Hits@10 &7.0 &2.4 &15.2 &4.3 &5.5 &8.2 &10.4 &6.4 &4.4 &3.4 &5.1 &1.1 &2.4 &4.2 &2.9 &1.6 \\ \cmidrule{2-19}
 & \multirow{2}{*}{NodePiece-QE} & Hits@3 &4.0 & &13.6 &3.2 &3.2 &3.4 &3.8 &2.4 &2.6 &1.7 &2.7 & & & & &  \\
 &  & Hits@10 &8.1 & &21.5 &6.4 &6.4 &7.8 &8.6 &5.9 &5.9 &3.8 &6.3 & & & & &  \\ \cmidrule{2-19}
 & \multirow{2}{*}{NodePiece-QE w/ GNN} & Hits@3 &18.1 & &39.1 &10.2 &3.8 &25.9 &31.0 &17.2 &13.8 &15.8 &6.4 & & & & & \\
 &  & Hits@10 &27.2 & &53.6 &16.0 &7.3 &39.2 &46.7 &26.2 &21.1 &24.0 &11.0 & & & & & \\ \cmidrule{2-19}
 & \multirow{2}{*}{GNN-QE} & Hits@3 &38.1 &22.7 &58.6 &24.5 &22.3 &53.0 &62.1 &39.0 &28.4 &33.4 &21.7 &26.4 &24.5 &13.6 &20.2 &28.5 \\
 &  & Hits@10 &49.4 &33.9 &71.7 &34.9 &31.3 &67.5 &74.7 &50.7 &38.9 &44.3 &30.9 &36.8 &39.5 &23.7 &29.6 &39.8 \\ \midrule
\multirow{8}{*}{106\%} & \multirow{2}{*}{Edge-type Heuristic} & Hits@3 &2.8 &1.3 &7.1 &1.9 &1.7 &3.5 &4.2 &1.8 &1.4 &1.4 &1.9 &1.6 &1.2 &1.2 &1.1 &1.4 \\
 &  & Hits@10 &6.4 &3.0 &14.5 &4.3 &4.4 &8.1 &9.7 &4.8 &3.7 &3.1 &4.7 &2.8 &2.2 &4.4 &2.6 &2.9 \\ \cmidrule{2-19}
 & \multirow{2}{*}{NodePiece-QE} & Hits@3 &4.0 & &11.9 &3.6 &4.0 &3.6 &4.0 &1.8 &2.1 &1.5 &3.3 & & & & & \\
 &  & Hits@10 &7.7 & &19.2 &5.9 &7.6 &8.1 &9.2 &4.3 &4.5 &4.1 &6.1 & & & & &  \\ \cmidrule{2-19}
 & \multirow{2}{*}{NodePiece-QE w/ GNN} & Hits@3 &22.1 & &39.6 &14.8 &5.1 &30.1 &35.6 &19.6 &19.6 &24.5 &9.9 & & & & & \\
 &  & Hits@10 &31.7 & &54.3 &21.6 &7.8 &44.2 &51.6 &28.4 &27.7 &35.2 &14.0 & & & & & \\ \cmidrule{2-19}
 & \multirow{2}{*}{GNN-QE} & Hits@3 &40.6 &28.3 &58.1 &28.5 &24.1 &54.7 &62.3 &38.7 &33.1 &40.3 &25.6 &31.7 &30.4 &17.0 &22.4 &40.1 \\
 &  & Hits@10 &50.4 &37.7 &71.9 &37.0 &32.4 &67.9 &73.7 &48.2 &41.8 &48.0 &32.7 &39.6 &43.2 &25.3 &31.1 &49.4 \\ \bottomrule

\end{tabular}
\end{adjustbox}
\end{table}

\begin{figure}[!ht]
    \centering
    \includegraphics[width=\textwidth]{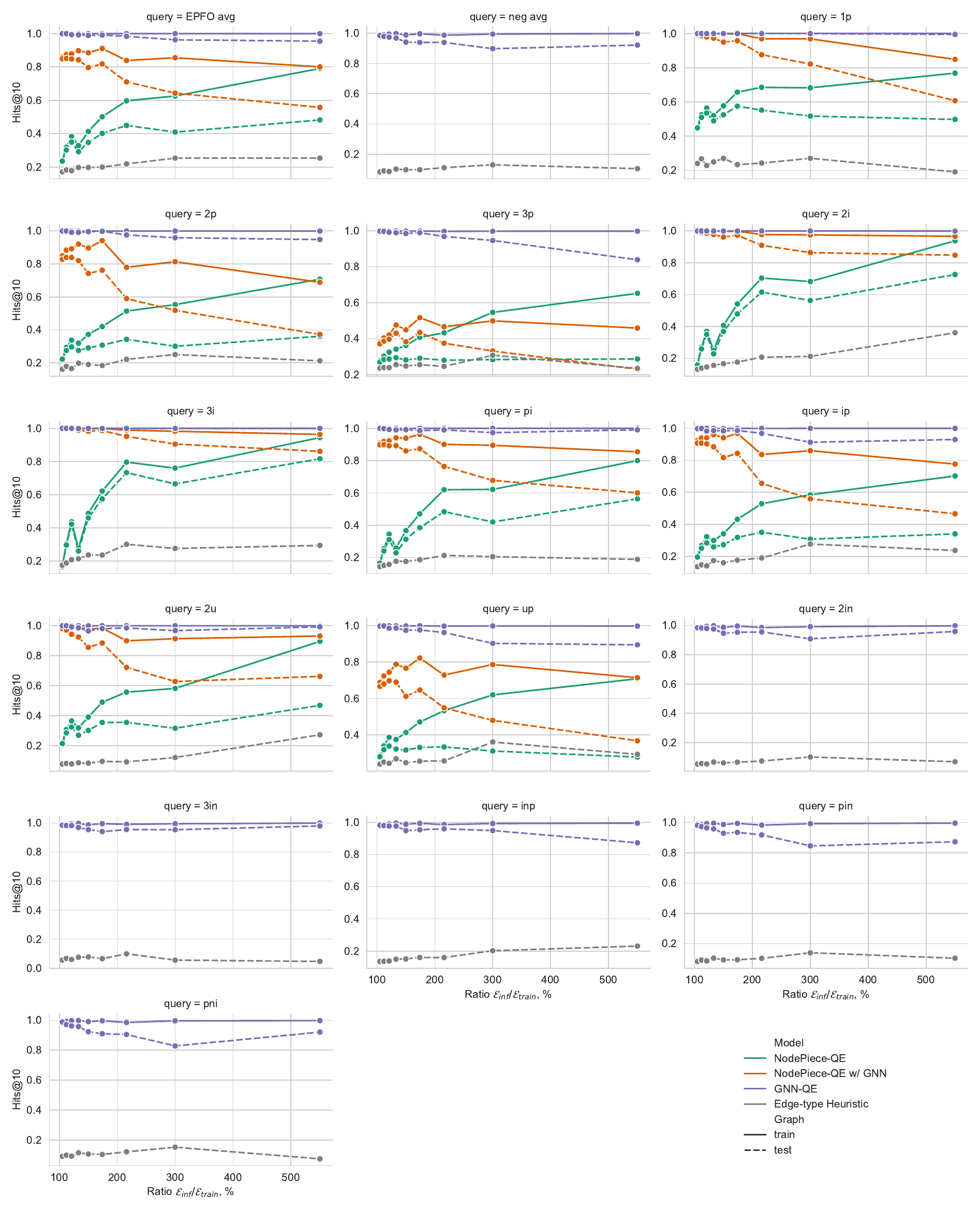}
    \caption{Hits@10 results on answering \textbf{training} queries executed over the original train (solid line) and test inference (dashed line) graphs. NodePiece-QE models are inference-only and were trained on \emph{1p} queries, GNN-QE is end-to-end trainable on all complex queries.}
    \label{fig:train_h10}
\end{figure}

Fig.~\ref{fig:train_h10} and Table~\ref{tab:more_train_queries} contain detailed results from the experiment in Section~\ref{sec:train_queries} about executing \textbf{training} queries over the original \emph{training} and extended \emph{test inference} graphs.

\begin{table}[t]
\caption{Hits@10 results (\%) of \textbf{training} queries executed over the original \emph{training} graph and extended \emph{test inference} graph. All ratios $\gE_{\textit{inf}} / \gE_{\textit{train}}$. avg$_p$ is the average on EPFO queries ($\land$, $\lor$). avg$_n$ is the average on queries with negation. NodePiece-QE models are inference-only and were trained on \emph{1p} queries, GNN-QE is end-to-end trainable on all complex queries.}
\label{tab:more_train_queries}
\begin{adjustbox}{width=\textwidth}
\begin{tabular}{@{}cllrrrrrrrrrrrrrrrr@{}}
\toprule
\multicolumn{1}{l}{Ratio} & Model & Graph & \bf{avg$_p$} & \bf{avg$_n$} & 1p & 2p & 3p & 2i & 3i & pi & ip & 2u & up & 2in & 3in & inp & pin & pni \\ \midrule
\multirow{7}{*}{550\%} & Edge-type Heuristic & test &25.4 &10.5 &19.2 &21.3 &23.4 &36.1 &29.4 &18.9 &23.6 &27.3 &29.2 &7.0 &4.6 &23.1 &10.4 &7.4 \\ \cmidrule{2-19}
 & \multirow{2}{*}{NodePiece-QE} & train &79.1 & &76.9 &70.7 &65.2 &93.8 &94.4 &80.0 &70.2 &89.6 &70.8 & & & & & \\
 &  & test &48.2 & &49.8 &36.1 &28.7 &72.7 &81.7 &56.4 &34.0 &46.9 &27.8 & & & & & \\ \cmidrule{2-19}
 & \multirow{2}{*}{NodePiece-QE w/ GNN} & train &80.0 & &84.9 &68.9 &45.9 &96.7 &96.3 &85.5 &77.6 &93.1 &71.4 & & & & & \\
 &  & test &55.7 & &60.8 &37.3 &23.0 &84.8 &86.1 &60.1 &46.6 &66.2 &36.7 & & & & &  \\ \cmidrule{2-19}
 & \multirow{2}{*}{GNN-QE} & train &99.9 &99.7 &100.0 &99.9 &99.8 &100.0 &100.0 &100.0 &100.0 &100.0 &99.8 &99.7 &100.0 &99.6 &99.6 &99.8 \\
 &  & test &95.4 &92.1 &99.5 &94.8 &83.9 &99.9 &100.0 &99.1 &93.0 &99.3 &89.5 &95.8 &98.1 &87.4 &87.3 &92.1 \\ \midrule
\multirow{7}{*}{300\%} & Edge-type Heuristic & test &25.4 &13.0 &27.1 &25.1 &30.8 &21.3 &27.6 &20.6 &27.7 &12.2 &36.0 &10.2 &5.5 &20.2 &14.0 &15.2 \\ \cmidrule{2-19}
 & \multirow{2}{*}{NodePiece-QE} & train &62.6 & &68.3 &55.3 &54.6 &68.3 &76.0 &62.2 &58.5 &58.2 &62.0 & & & & & \\
 &  & test &41.0 & &51.8 &30.1 &28.4 &56.3 &66.5 &42.1 &30.7 &31.7 &31.1 & & & & & \\ \cmidrule{2-19}
 & \multirow{2}{*}{NodePiece-QE w/ GNN} & train &85.5 & &96.9 &81.3 &49.9 &97.5 &98.2 &89.5 &86.0 &91.4 &78.7 & & & & & \\
 &  & test &64.3 & &82.2 &51.9 &33.1 &86.4 &90.5 &67.8 &55.9 &62.8 &47.9 & & & & & \\ \cmidrule{2-19}
 & \multirow{2}{*}{GNN-QE} & train &99.9 &99.4 &100.0 &100.0 &99.8 &100.0 &100.0 &100.0 &100.0 &100.0 &99.8 &99.1 &99.6 &99.4 &99.2 &99.6 \\
 &  & test &96.2 &89.7 &100.0 &95.9 &94.6 &100.0 &100.0 &97.3 &91.3 &96.7 &90.3 &90.8 &95.5 &95.0 &84.6 &82.8 \\ \midrule
\multirow{7}{*}{217\%} & Edge-type Heuristic & test &21.9 &11.1 &24.3 &22.2 &24.6 &20.8 &30.0 &21.4 &18.9 &9.3 &25.6 &7.4 &9.9 &16.0 &10.3 &12.1 \\ \cmidrule{2-19}
 & \multirow{2}{*}{NodePiece-QE} & train &59.7 & &68.6 &51.4 &43.3 &70.4 &79.6 &62.0 &52.8 &55.8 &53.3 & & & & & \\
 &  & test &45.0 & &55.3 &34.3 &28.0 &61.6 &73.3 &48.4 &35.0 &35.6 &33.3 & & & & & \\ \cmidrule{2-19}
 & \multirow{2}{*}{NodePiece-QE w/ GNN} & train &83.9 & &96.9 &77.9 &46.6 &97.8 &98.9 &90.1 &83.6 &90.0 &72.9 & & & & & \\
 &  & test &71.0 & &87.6 &59.0 &37.5 &91.0 &95.2 &76.4 &65.6 &72.2 &54.9 & & & & & \\ \cmidrule{2-19}
 & \multirow{2}{*}{GNN-QE} & train &99.9 &98.7 &100.0 &99.9 &99.7 &100.0 &100.0 &100.0 &99.9 &100.0 &99.8 &98.6 &99.2 &98.7 &98.3 &98.5 \\
 &  & test &98.3 &93.9 &100.0 &97.6 &96.9 &100.0 &100.0 &99.0 &96.8 &98.5 &96.2 &95.5 &95.6 &96.1 &91.8 &90.5 \\ \midrule
\multirow{7}{*}{175\%} & Edge-type Heuristic & test &20.0 &9.8 &23.4 &18.4 &25.5 &17.7 &23.6 &18.7 &17.6 &9.6 &25.5 &6.7 &6.5 &16.0 &9.3 &10.3 \\ \cmidrule{2-19}
 & \multirow{2}{*}{NodePiece-QE} & train &50.1 & &65.8 &42.0 &40.7 &54.2 &62.1 &47.1 &43.1 &49.0 &47.0 & & & & & \\
 &  & test &40.2 & &57.6 &30.8 &29.1 &47.9 &57.4 &38.6 &31.8 &35.5 &33.0 & & & & & \\ \cmidrule{2-19}
 & \multirow{2}{*}{NodePiece-QE w/ GNN} & train &91.0 & &99.9 &94.1 &51.7 &99.9 &99.9 &96.3 &96.8 &98.4 &82.2 & & & & & \\
 &  & test &81.8 & &95.8 &76.2 &43.4 &97.3 &98.7 &87.4 &84.3 &88.5 &64.6 & & & & & \\ \cmidrule{2-19}
 & \multirow{2}{*}{GNN-QE} & train &100.0 &99.6 &100.0 &100.0 &99.9 &100.0 &100.0 &100.0 &100.0 &100.0 &99.9 &99.6 &99.6 &99.5 &99.4 &99.6 \\
 &  & test &99.1 &93.9 &100.0 &99.8 &98.9 &100.0 &99.9 &98.6 &98.7 &98.1 &97.6 &95.3 &94.1 &95.5 &93.5 &91.0 \\ \midrule
\multirow{7}{*}{150\%} & Edge-type Heuristic & test &19.7 &9.8 &27.0 &19.1 &24.8 &16.6 &23.6 &17.6 &15.9 &8.4 &24.7 &6.2 &7.7 &15.1 &9.2 &10.6 \\ \cmidrule{2-19}
 & \multirow{2}{*}{NodePiece-QE} & train &41.3 & &57.8 &37.3 &36.3 &40.7 &48.7 &36.8 &34.0 &39.1 &41.3 & & & & & \\
 &  & test &34.8 & &52.5 &29.1 &28.1 &36.9 &46.0 &31.4 &27.2 &30.2 &31.6 & & & & & \\ \cmidrule{2-19}
 & \multirow{2}{*}{NodePiece-QE w/ GNN} & train &88.5 & &99.7 &89.7 &44.9 &99.6 &99.8 &93.9 &94.1 &97.7 &76.6 & & & & & \\
 &  & test &79.6 & &95.0 &74.3 &38.3 &96.2 &98.3 &86.0 &81.7 &85.6 &61.1 & & & & &  \\ \cmidrule{2-19}
 & \multirow{2}{*}{GNN-QE} & train &99.9 &98.8 &100.0 &99.9 &99.8 &100.0 &100.0 &99.9 &99.9 &100.0 &99.9 &98.6 &98.8 &98.9 &98.7 &99.1 \\
 &  & test &98.8 &94.1 &100.0 &99.6 &98.5 &99.9 &99.9 &99.2 &98.4 &96.5 &97.4 &94.6 &95.5 &94.9 &92.9 &92.3 \\ \midrule
\multirow{7}{*}{133\%} & Edge-type Heuristic & test &19.8 &10.2 &25.0 &19.8 &25.5 &15.6 &21.3 &17.8 &17.3 &8.7 &26.8 &6.8 &7.5 &15.0 &10.4 &11.5 \\ \cmidrule{2-19}
 & \multirow{2}{*}{NodePiece-QE} & train &32.9 & &52.0 &32.0 &34.1 &25.0 &27.7 &26.0 &29.9 &31.9 &37.4 & & & & & \\
 &  & test &29.2 & &49.0 &27.5 &29.4 &22.9 &26.0 &23.0 &25.8 &27.0 &32.1 & & & & &  \\ \cmidrule{2-19}
 & \multirow{2}{*}{NodePiece-QE w/ GNN} & train &89.7 & &99.9 &92.0 &47.6 &99.9 &99.9 &94.1 &96.0 &99.2 &78.8 & & & & & \\
 &  & test &84.2 & &97.3 &82.0 &43.0 &97.8 &98.9 &89.3 &88.5 &92.4 &69.0 & & & & & \\ \cmidrule{2-19}
 & \multirow{2}{*}{GNN-QE} & train &100.0 &99.7 &100.0 &100.0 &99.9 &100.0 &100.0 &100.0 &100.0 &100.0 &99.9 &99.8 &99.8 &99.6 &99.5 &99.8 \\
 &  & test &99.2 &96.8 &100.0 &99.1 &98.9 &99.9 &99.8 &99.0 &98.5 &98.6 &98.7 &97.5 &97.0 &97.7 &95.8 &95.8  \\ \midrule
\multirow{7}{*}{121\%} & Edge-type Heuristic & test &17.9 &8.6 &22.9 &16.6 &23.9 &14.6 &20.7 &15.8 &14.0 &7.9 &24.3 &5.4 &5.9 &13.8 &8.6 &9.2 \\ \cmidrule{2-19}
 & \multirow{2}{*}{NodePiece-QE} & train &38.4 & &56.4 &33.8 &32.5 &37.0 &43.6 &34.5 &32.3 &36.6 &38.6 & & & & &  \\
 &  & test &35.0 & &53.5 &29.7 &28.7 &35.1 &42.2 &31.2 &28.4 &32.4 &33.8 & & & & &  \\ \cmidrule{2-19}
 & \multirow{2}{*}{NodePiece-QE w/ GNN} & train &87.8 & &100.0 &89.2 &42.1 &99.9 &99.9 &92.3 &94.1 &98.5 &74.4 & & & & & \\
 &  & test &84.8 & &98.0 &83.9 &39.6 &98.6 &99.4 &89.2 &90.3 &94.3 &69.7 & & & & & \\ \cmidrule{2-19}
 & \multirow{2}{*}{GNN-QE} & train  &100.0 &99.4 &100.0 &100.0 &99.9 &100.0 &100.0 &100.0 &100.0 &100.0 &99.9 &99.4 &99.5 &99.1 &99.3 &99.7 \\
 &  & test &99.3 &97.3 &100.0 &99.2 &99.2 &100.0 &100.0 &99.3 &98.3 &99.2 &98.6 &97.9 &98.3 &97.8 &96.5 &96.2 \\ \midrule
\multirow{7}{*}{113\%} & Edge-type Heuristic & test &18.3 &9.0 &26.8 &17.9 &23.9 &13.9 &18.8 &15.2 &14.6 &8.2 &25.0 &5.7 &6.7 &13.6 &9.1 &9.8 \\ \cmidrule{2-19}
 & \multirow{2}{*}{NodePiece-QE} & train &31.8 & &52.5 &29.4 &30.3 &27.0 &30.5 &25.5 &26.7 &30.8 &33.9 & & & & &  \\
 &  & test &30.2 & &51.0 &27.6 &28.4 &25.9 &29.6 &24.1 &25.0 &28.6 &31.8 & & & & &  \\ \cmidrule{2-19}
 & \multirow{2}{*}{NodePiece-QE w/ GNN} & train &87.4 & &100.0 &88.4 &40.4 &99.9 &99.9 &91.9 &94.0 &99.6 &72.4 & & & & & \\
 &  & test &85.1 & &98.8 &83.9 &38.6 &99.0 &99.5 &89.9 &90.7 &97.2 &67.9 & & & & & \\ \cmidrule{2-19}
 & \multirow{2}{*}{GNN-QE} & train &100.0 &98.8 &100.0 &100.0 &99.9 &100.0 &100.0 &100.0 &100.0 &100.0 &100.0 &98.7 &99.0 &98.7 &98.5 &99.2 \\
 &  & test &99.9 &97.8 &100.0 &100.0 &99.7 &100.0 &100.0 &99.8 &99.9 &100.0 &99.9 &98.2 &98.4 &98.1 &97.3 &97.1 \\ \midrule
\multirow{7}{*}{106\%} & Edge-type Heuristic & test &17.1 &8.3 &24.1 &16.1 &23.6 &13.2 &17.4 &14.4 &13.5 &7.8 &23.9 &5.4 &5.5 &13.5 &8.2 &9.0 \\ \cmidrule{2-19}
 & \multirow{2}{*}{NodePiece-QE} & train &24.1 & &45.4 &22.9 &27.6 &16.1 &16.8 &17.0 &20.1 &22.1 &28.7 & & & & &  \\
 &  & test &23.6 & &44.9 &22.3 &27.0 &15.8 &16.6 &16.6 &19.5 &21.5 &27.9 & & & & & \\ \cmidrule{2-19}
 & \multirow{2}{*}{NodePiece-QE w/ GNN} & train &86.0 & &99.9 &84.9 &38.0 &99.9 &99.9 &90.8 &92.3 &99.2 &68.8 & & & & & \\
 &  & test &84.9 & &99.4 &82.8 &37.1 &99.5 &99.8 &89.8 &90.7 &98.1 &66.6 & & & & & \\ \cmidrule{2-19}
 & \multirow{2}{*}{GNN-QE} & train &100.0 &99.0 &100.0 &100.0 &99.9 &100.0 &100.0 &100.0 &99.9 &100.0 &99.9 &98.8 &99.0 &98.9 &98.9 &99.4 \\
 &  & test &99.9 &98.4 &100.0 &100.0 &99.8 &100.0 &100.0 &100.0 &99.8 &100.0 &99.7 &98.4 &98.5 &98.3 &98.1 &98.8 \\ \bottomrule

\end{tabular}
\end{adjustbox}
\end{table}

\section{Hyperparameters}
\label{app:hyperparams}

Both NodePiece-QE and GNN-QE models are implemented with PyTorch~\cite{DBLP:conf/nips/PaszkeGMLBCKLGA19} (MIT License).
In particular, NodePiece-QE models employ PyG~\cite{Fey/Lenssen/2019} (MIT License) and PyKEEN~\cite{ali2021pykeen} (MIT License) for training link prediction models.
GNN-QE is implemented based on the official NBFNet repository~\footnote{\url{https://github.com/DeepGraphLearning/NBFNet}} (MIT License) and TorchDrug~\cite{zhu2022torchdrug} library (Apache 2.0).

For all inductive experiments in Sections~\ref{sec:test_queries} and \ref{sec:train_queries}, Table~\ref{tab:nodepieceqe_hparam} lists best hyperparameters for NodePiece-QE models without GNN encoder, Table~\ref{tab:nodepieceqe_gnn_hparam} contains hyperparameters for GNN-enabled NodePiece-QE models. 
The GNN-free models use only relation-based tokenization where each entity $e$ is represented with two fixed-size sets: a set of $k$ unique \emph{incoming} $r_i$ and a set of $k$ unique \emph{outgoing} $r_o$ relation types. 
Looking up their $d$-dimensional vectors, we obtain:

\begin{align*}
    e = \Big[ [\vr_{i1}, \vr_{i2}, \dots, \vr_{ik}] [\vr_{o1}, \vr_{o2}, \dots, \vr_{ok}] \Big] \in \sR^{2 \times k \times d}
\end{align*}

If, for some entity, the number of unique relations of a certain kind is less than $k$, we pad the set with auxiliary \texttt{[PAD]} tokens.
Entity representations are built as a function of the two sets $f(e): \sR^{2 \times k \times d} \rightarrow \sR^d$:
\begin{align*}
    \vh_e = \textsc{MLP}\Big( \textsc{RandomProj} (\sum_{j=0}^{k} \vr_{ij}) + \textsc{RandomProj}(\sum_{j=0}^{k} \vr_{oj}) \Big)
\end{align*}

Particularly, we first sum up tokens of the same direction, pass them through a random projection layer $\textsc{RandomProj}$ (we found that making this projection learnable does not improve results), sum up representations of \emph{incoming} and \emph{outgoing} parts, and pass the resulting vector through a learnable MLP. 
This way, the number of learnable encoder parameters does not depend on the sequence length $k$, i.e., the number of chosen tokens per node.

The GNN-enabled models employ a slightly different \emph{Concat + MLP} encoder where each node is tokenized with a sample of $k$ incident relations. Then, we concatenate $d$-dimensional embeddings of those \emph{tokens} $t_i$ into a single long vector $\sR^{kd}$, and then use a 2-layer MLP to project it to a model dimension $d$, i.e., $f(e): \sR^{kd} \rightarrow \sR^d$:

\begin{align*}
    \vh_e = \textsc{MLP}\Big( [\vt_0; \vt_1; \dots; \vt_k] \Big)
\end{align*}

For the large-scale experiment on WikiKG-QE in Section~\ref{sec:scaling}, we employ the \emph{Concat + MLP} encoder. Instead of separating incoming and outgoing relation types, we first tokenize each node with 20 nearest \emph{anchors} (pre-selected in advance using the default NodePiece strategy~\cite{galkin2022nodepiece}) and add a sample of $k$ unique incident relations. 
NodePiece-QE w/ GNN employs a 3-layer CompGCN with the RotatE interaction function during message computation and \emph{sum} aggregator. Due to high memory consumption, we train the 50d model for 2 epochs on 2 x RTX 8000 (48 GB) GPUs.

The overall tokens vocabulary consists of 20,000 anchor nodes, 1,024 relation types (including inverse relations) and one \texttt{[PAD]} token. 
All hyperparameters for this experiments are listed in Table~\ref{tab:nodepieceqe_wikikg}.

Having trained the link predictors, we tune CQD-Beam hyperparameters on the validation set  varying the t-norms, t-conorms, and scores normalization. 
Table~\ref{tab:cqd_hparams} lists best options for each EPFO query type. 
For all experiments, we used a beam size 
%$k=32$ except for \emph{3p} queries on WikiKG-QE where we used $k=8$ 
$k=32$ except for queries on WikiKG-QE where we used $k=8$ 
due to the memory-expensive need of maintaining a beam over 2M entities.

Table~\ref{tab:nbfnetqe_hyperparameter} lists hyperparameters for GNN-QE models for all inductive splits. 
We found this architecture is quite stable under various configurations and eventually employed the same set of hyperparameters across all datasets.

\begin{table}[ht]
\centering
\caption{NodePiece-QE hyperparameters for all inductive splits.}
\label{tab:nodepieceqe_hparam}
%\scriptsize
%\begin{adjustbox}{width=\textwidth}
\begin{tabular}{lccccccccc}\toprule
\multirow{2}{*}{Hyperparameter} &\multicolumn{9}{c}{\textbf{Dataset $\gE_{\textit{inf}} / \gE_{\textit{train}}$ Ratios}} \\\cmidrule{2-10}
&\textbf{106} &\textbf{113} &\textbf{133} &\textbf{134} &\textbf{150} &\textbf{175} &\textbf{217} &\textbf{300} &\textbf{550} \\\midrule
Vocab size &472 &466 &460 &458 &450 &438 &442 &402 &346 \\
Tokens per node & 20 & 20 & 20 & 20 & 20 & 20 & 20 & 20 & 10 \\
Vocab dim &400 &400 &400 &400 &400 &400 &400 &400 &  1000 \\
Scoring function &\multicolumn{9}{c}{ComplEx~\cite{trouillon2016complex}} \\ \midrule
Encoder &\multicolumn{9}{c}{RandomProj + MLP} \\
Encoder dim &400 &400 &400 &400 &400 &400 &400 &400 & 1000 \\
Encoder layers &\multicolumn{9}{c}{2} \\ \midrule
Batch size &\multicolumn{9}{c}{256} \\
Epochs & 400 & 400 & 400 & 1000 & 1000 & 2000 & 2000 & 2000 & 3000 \\
Learning rate &\multicolumn{9}{c}{1e-4} \\
Optimizer &\multicolumn{9}{c}{Adam} \\
Loss function &\multicolumn{9}{c}{BCE} \\
Adv. temperature & 1.0 & 1.0 & 1.0 & 1.0 & 0.5 & 0.5 & 0.5 & 0.2 & 1.0 \\
%Margin &3 &3 &3 &10 &3 &10 &3 &3 &3 \\
\# negatives & 128 & 128 & 128 & 128 & 128 & 128 & 128 & 128 & 128 \\ \midrule
\# parameters &699k &694k &689k &688k &681k &671k &675k &643k &2.7M \\
Training time (hrs) & 15 & 12 & 10 & 9 & 6 & 9 & 9 & 5 & 1 \\
\bottomrule
\end{tabular}
%\end{adjustbox}
\end{table}

\begin{table}[ht]
\centering
\caption{NodePiece-QE with GNN hyperparameters for all inductive splits.}
\label{tab:nodepieceqe_gnn_hparam}
%\scriptsize
\begin{tabular}{lccccccccc}\toprule
\multirow{2}{*}{Hyperparameter} &\multicolumn{9}{c}{\textbf{Dataset $\gE_{\textit{inf}} / \gE_{\textit{train}}$ Ratios}} \\\cmidrule{2-10}
&\textbf{106} &\textbf{113} &\textbf{133} &\textbf{134} &\textbf{150} &\textbf{175} &\textbf{217} &\textbf{300} &\textbf{550} \\\midrule
Vocab size &472 &466 &460 &458 &450 &438 &442 &402 &346 \\
Tokens per node &\multicolumn{8}{c}{20} & 10 \\
Vocab dim &\multicolumn{8}{c}{200} & 400 \\
Scoring function &\multicolumn{9}{c}{ComplEx~\cite{trouillon2016complex}} \\ \midrule
Encoder &\multicolumn{9}{c}{Concat + MLP} \\
Encoder dim &\multicolumn{8}{c}{200} & 400 \\
Encoder layers &\multicolumn{9}{c}{2} \\ \midrule
GNN encoder &\multicolumn{9}{c}{CompGCN~\cite{Vashishth2020Composition-based} + RotatE~\cite{DBLP:conf/iclr/SunDNT19} message function} \\
GNN layers & 5 & 5 & 5 & 5 & 5 & 3 & 3 & 3 & 3 \\
GNN dim &\multicolumn{8}{c}{200} & 400 \\ \midrule
% GNN attn dropout &\multicolumn{9}{c}{0.1} \\
% GNN attn heads &\multicolumn{9}{c}{2} \\ \midrule
Batch size &\multicolumn{9}{c}{256} \\
Epochs & 600 & 1000 & 1000 & 1000 & 1000 & 1000 & 3000 & 4000 & 4000 \\
Learning rate &\multicolumn{9}{c}{5e-4} \\
Optimizer &\multicolumn{9}{c}{Adam} \\
Loss function &\multicolumn{9}{c}{BCE} \\
Adv. temperature & 1.0 & 1.0 & 1.0 & 1.0 & 1.0 & 1.0 & 1.0 & 0.5 & 1.0 \\
%Margin &20 &20 &20 &20 &20 &20 &20 &20 &5 \\
\# negatives & 128 & 128 & 128 & 128 & 128 & 128 & 128 & 128 & 128 \\  \midrule
\# parameters & 2.8M & 2.8M & 2.8M & 2.8M & 2.8M & 2.4M & 2.4M & 2.3M & 5.7M \\
Training time (hrs) & 30 & 30 & 19 & 20 & 14 & 15 & 8 & 5 & 1 \\
\bottomrule
\end{tabular}
\end{table}

\begin{table}[ht]
\centering
\caption{NodePiece-QE hyperparameters for WikiKG-QE (133\%).}
\label{tab:nodepieceqe_wikikg}
%\scriptsize
%\begin{adjustbox}{width=\textwidth}
\begin{tabular}{lcc}\toprule
Hyperparameter & \textbf{NodePiece-QE} & \textbf{NodePiece-QE w/ GNN} \\ \midrule
Vocab size & \multicolumn{2}{c}{20,000 anchors + 1024 relation types}  \\
Anchor tokens per node & 20 & 15 \\
Relation tokens per node & 20 & 10 \\ 
Vocab dim & 100 & 50  \\
Scoring function & \multicolumn{2}{c}{ComplEx~\cite{trouillon2016complex}} \\ \midrule
Encoder & \multicolumn{2}{c}{Concat + MLP} \\
Encoder dim & 200 & 100 \\
Encoder layers & 2 & 2 \\ \midrule
GNN & - & CompGCN \\ 
GNN dim & - & 50 \\ 
GNN layers & - & 3 \\ \midrule
Batch size & 512 & 1024 \\
% Epochs & 60 ($\approx$ 1.5M steps) \\
Epochs & 40 ($\approx$ 1M steps) & 2 \\
Learning rate & \multicolumn{2}{c}{1e-4} \\
Optimizer &  \multicolumn{2}{c}{Adam} \\
%Loss function & Negative Sampling Self-adversarial Loss~\cite{DBLP:conf/iclr/SunDNT19} \\
%Margin & 30 \\
Loss function & \multicolumn{2}{c}{BCE} \\
Adversarial temp. & \multicolumn{2}{c}{1.0} \\
\# negatives & 64 & 512 \\ \midrule
%\# parameters & 2,925,300 \\
\# parameters & 2,922,900 & 1,211,950 \\
Training time (hrs) & 40 & 16 \\
\bottomrule
\end{tabular}
%\end{adjustbox}
\end{table}

\begin{table}[!htp]
\centering
\caption{CQD-Beam t-norm hyperparameters for all splits and both link predictors, NodePiece-QE and NodePiece-QE w/ GNN, when answering EPFO queries. The default beam size $k = 32$, \emph{prod} + $\sigma$ is product t-norm with sigmoid score normalization. Details on t-norms are in Appendix~\ref{app:tnorms}.}
\label{tab:cqd_hparams}
%\scriptsize
\begin{adjustbox}{width=\textwidth}
\begin{tabular}{llcccccccccc}
\toprule
Ratio &Link predictor &\textbf{1p} &\textbf{2p} &\textbf{3p} &\textbf{2i} &\textbf{3i} &\textbf{pi} &\textbf{ip} &\textbf{2u} &\textbf{up} \\ \midrule
\multirow{2}{*}{106\%} &NodePiece-QE &  \multirow{2}{*}{prod + $\sigma$} &  \multirow{2}{*}{prod + $\sigma$} &  \multirow{2}{*}{prod + $\sigma$} &\multirow{2}{*}{prod + $\sigma$} &\multirow{2}{*}{prod + $\sigma$} & \multirow{2}{*}{prod + $\sigma$} &\multirow{2}{*}{prod + $\sigma$} &\multirow{2}{*}{prod + $\sigma$} &\multirow{2}{*}{prod + $\sigma$} \\ 
&NodePiece-QE w/ GNN & & & & & & & & & \\ \midrule
\multirow{2}{*}{113\%} &NodePiece-QE &\multirow{2}{*}{prod + $\sigma$} &\multirow{2}{*}{prod + $\sigma$} &\multirow{2}{*}{prod + $\sigma$} &\multirow{2}{*}{prod + $\sigma$} &\multirow{2}{*}{prod + $\sigma$} &\multirow{2}{*}{prod + $\sigma$} &\multirow{2}{*}{prod + $\sigma$} &\multirow{2}{*}{prod + $\sigma$} &\multirow{2}{*}{prod + $\sigma$} \\
&NodePiece-QE w/ GNN & & & & & & & & & \\ \midrule
\multirow{2}{*}{122\%} &NodePiece-QE &  \multirow{2}{*}{prod + $\sigma$} &  \multirow{2}{*}{prod + $\sigma$} & \multirow{2}{*}{prod + $\sigma$} &\multirow{2}{*}{prod + $\sigma$} &\multirow{2}{*}{prod + $\sigma$} &\multirow{2}{*}{prod + $\sigma$} &  \multirow{2}{*}{prod + $\sigma$} &\multirow{2}{*}{prod + $\sigma$} &\multirow{2}{*}{prod + $\sigma$} \\
&NodePiece-QE w/ GNN & & & & & & & & & \\ \midrule
\multirow{2}{*}{134\%} &NodePiece-QE & \multirow{2}{*}{prod + $\sigma$} &\multirow{2}{*}{prod + $\sigma$} &\multirow{2}{*}{prod + $\sigma$} &\multirow{2}{*}{prod + $\sigma$} &\multirow{2}{*}{prod + $\sigma$} &\multirow{2}{*}{prod + $\sigma$} &\multirow{2}{*}{prod + $\sigma$} &\multirow{2}{*}{prod + $\sigma$} &\multirow{2}{*}{prod + $\sigma$} \\
&NodePiece-QE w/ GNN & & & & & & & & & \\ \midrule
\multirow{2}{*}{150\%} &NodePiece-QE & \multirow{2}{*}{prod + $\sigma$} &\multirow{2}{*}{prod + $\sigma$} &\multirow{2}{*}{prod + $\sigma$} &\multirow{2}{*}{prod + $\sigma$} &\multirow{2}{*}{prod + $\sigma$} &\multirow{2}{*}{prod + $\sigma$} &\multirow{2}{*}{prod + $\sigma$} &\multirow{2}{*}{prod + $\sigma$} &\multirow{2}{*}{prod + $\sigma$} \\
&NodePiece-QE w/ GNN & & & & & & & & & \\ \midrule
\multirow{2}{*}{175\%} &NodePiece-QE &\multirow{2}{*}{prod + $\sigma$} & \multirow{2}{*}{prod + $\sigma$} & \multirow{2}{*}{prod + $\sigma$} &\multirow{2}{*}{prod + $\sigma$} &\multirow{2}{*}{prod + $\sigma$} &\multirow{2}{*}{prod + $\sigma$} &\multirow{2}{*}{prod + $\sigma$} & \multirow{2}{*}{prod + $\sigma$} &\multirow{2}{*}{prod + $\sigma$} \\
&NodePiece-QE w/ GNN & & & & & & & & & \\ \midrule
\multirow{2}{*}{217\%} &NodePiece-QE & \multirow{2}{*}{prod + $\sigma$} &\multirow{2}{*}{prod + $\sigma$} &\multirow{2}{*}{prod + $\sigma$} &\multirow{2}{*}{prod + $\sigma$} &\multirow{2}{*}{prod + $\sigma$} &\multirow{2}{*}{prod + $\sigma$} &\multirow{2}{*}{prod + $\sigma$} &\multirow{2}{*}{prod + $\sigma$} & \multirow{2}{*}{prod + $\sigma$} \\
&NodePiece-QE w/ GNN &  & & & & & & & & \\ \midrule
\multirow{2}{*}{300\%} &NodePiece-QE & \multirow{2}{*}{prod + $\sigma$} & \multirow{2}{*}{prod + $\sigma$} &\multirow{2}{*}{prod + $\sigma$} &\multirow{2}{*}{prod + $\sigma$} &\multirow{2}{*}{prod + $\sigma$} &\multirow{2}{*}{prod + $\sigma$} &\multirow{2}{*}{prod + $\sigma$} &\multirow{2}{*}{prod + $\sigma$} &\multirow{2}{*}{prod + $\sigma$} \\
&NodePiece-QE w/ GNN & & & & & & & & & \\ \midrule
\multirow{2}{*}{550\%} &NodePiece-QE &\multirow{2}{*}{prod + $\sigma$} & \multirow{2}{*}{prod + $\sigma$} & \multirow{2}{*}{prod + $\sigma$} & \multirow{2}{*}{prod + $\sigma$} & \multirow{2}{*}{prod + $\sigma$} & \multirow{2}{*}{prod + $\sigma$} & \multirow{2}{*}{prod + $\sigma$} & \multirow{2}{*}{prod + $\sigma$} & \multirow{2}{*}{prod + $\sigma$} \\
&NodePiece-QE w/ GNN & & & & & & & & & \\
\midrule 
\midrule
133\% & NodePiece-QE & \multicolumn{9}{c}{ (WikiKG-QE) prod + $\sigma$ for all query types} \\
\bottomrule
\end{tabular}
\end{adjustbox}
\end{table}

BetaE (as a transductive baseline for the reference 175\% dataset) was configured with $400d$ embedding dimension, batch size 512, 32 negative samples, learning rate 0.0005, margin $\gamma$ 60, and trained on 10 query patterns \emph{\{1p,2p,3p,2i,3i,2in,3in,inp,pin,pni\}} for $200k$ steps.

\clearpage
\begin{table*}[!h]
    \centering
    \caption{Hyperparameters of GNN-QE on different datasets. All the hyperparameters are selected by the performance on the validation set.}
    \footnotesize
    \begin{tabular}{llc}
        \toprule
        \multicolumn{2}{l}{\bf{Hyperparameter}} & \bf{All splits}\\
        \midrule
        \multirow{4}{*}{\bf{GNN}}
        & \#layers & 4 \\
        & hidden dim. & 32 \\
        & composition & DistMult~\cite{DBLP:journals/corr/YangYHGD14} \\
        & aggregation & PNA~\cite{corso2020principal} \\
        \midrule
        \multirow{2}{*}{\bf{MLP}}
        & \#layer & 2 \\
        & hidden dim. & 64  \\
        \midrule
        \bf{Traversal Dropout} & probability & 0.5 \\
        \bf{Logical Operator} & t-norm & product \\
        \midrule
        \multirow{8}{*}{\bf{Learning}}
        & batch size & 64  \\
        & sample weight & uniform across queries  \\
        & loss & BCE \\
        & \# negatives & 32 \\
        & optimizer        & Adam  \\
        & learning rate    & 5e-3  \\
        & iterations (\#batch) & 10,000  \\
        & adv. temperature & 0.1  \\
        \bottomrule
    \end{tabular}
    \label{tab:nbfnetqe_hyperparameter}
\end{table*}

% \section{Identifying Easy and Hard Answers}
% \label{app:easy_vs_hard}

% \edit{
% The idea of computing ROC AUC is as follows:
% Suppose we have a list of unfiltered raw scores from which we extract scores of all easy and hard answers. Suppose we have a query with 4 easy and 1 hard answer: $[5, 6, 7, 8, 32]$  where 8 is a rank of a hard answer while 5, 6, 7, 32 are ranks of easy answers. 
% We then create binary labels for the scores assigning 1 to the hard answers, e.g., $[0, 0, 0, 1, 0]$.}

% \edit{
% Given those two arrays, we then compute the ROC AUC score that would measure how many hard answers are ranked after easy answers, e.g., in our example ROC AUC is 0.75. 
% Note that the score  does not depend on actual values of ranks, that is, the metric will be high when easy answers are, e.g., ranked 1000-1004 as long as hard answers are ranked 1005 and lower. 
% Therefore, ROC AUC still needs to be paired with MRR to see where easy and hard answers are ranked absolutely.}

\section{Edge-type Heuristic}
\label{app:dummy}

We consider \emph{Edge-type Heuristic} as a trivial baseline for inductive complex query. Given a query $\gQ = (\gC, \gR_Q, \gG)$, Edge-type Heuristic finds all entities $e \in \gE$ that satisfy the relations in the last hop of $\gR_Q$ on the inference graph $\gG_{\textit{inf}}$. In other words, this baseline filters out entities that are not consistent with the query according to the edge types, which is a necessary condition for the answers when the inference graph is reasonably dense. Since Edge-type Heuristic only distinguishes the entities into two classes, we randomly shuffle the entities in each class to create a ranking.

Edge-type Heuristic can be easily implemented as GNN-QE with a special relation \emph{projection}. Given an inference graph $\gG_{\textit{inf}}$, we first preprocess a relation-to-entity mapping $\mM$, where $\mM_{r, t}$ means there exists a head entity $h$ and an edge $(h, r, t)$ for tail entity $t$. Then the relation \emph{projection} of relation $r$ can be implemented by looking up the row $\mM_r$. Note that the Edge-type Heuristic only filters entities according to the edge type, and hence the head entity (or the distribution of head entity) is ignored in the relation projection.

\end{document}